\documentclass[a4paper,fleqn]{cas-dc}

\usepackage[authoryear,longnamesfirst]{natbib}
\usepackage{comment}
\def\tsc#1{\csdef{#1}{\textsc{\lowercase{#1}}\xspace}}
\tsc{WGM}
\tsc{QE}
\usepackage{tikz}

\usepackage{placeins}
\usepackage{algorithm}
\usepackage{algpseudocode}

\usepackage{tcolorbox}
\newtcolorbox{todo}{
  colback=white!10,
  colframe=red,
  title=Todo
}

\begin{document}
\let\WriteBookmarks\relax
\def\floatpagepagefraction{1}
\def\textpagefraction{.001}

% Short title
\shorttitle{}    

% Short author
\shortauthors{}  

% Main title of the paper
\title [mode = title]{Robust Fault Detection in Mechanical Multimodal Time Series via Self-Supervised Cross-Modal Reconstruction}  

% Title footnote mark
% eg: \tnotemark[1]
%\tnotemark[1] 

% Title footnote 1.
% eg: \tnotetext[1]{Title footnote text}
%\tnotetext[1]{} 

% First author
%
% Options: Use if required
% eg: \author[1,3]{Author Name}[type=editor,
%       style=chinese,
%       auid=000,
%       bioid=1,
%       prefix=Sir,
%       orcid=0000-0000-0000-0000,
%       facebook=<facebook id>,
%       twitter=<twitter id>,
%       linkedin=<linkedin id>,
%       gplus=<gplus id>]

\author[1]{Magnus Munk Jensen}[ 
orcid = 0009-0008-3749-547X
]

% Corresponding author indication
\cormark[1]
% Corresponding author text
\cortext[1]{Corresponding author}

% Footnote of the first author
%\fnmark[1]

% Email id of the first author
\ead{magnusmj@es.aau.dk}

% URL of the first author
%\ead[url]{https://vbn.aau.dk/da/persons/magnusmj/}

% Credit authorship
% eg: \credit{Conceptualization of this study, Methodology, Software}
\credit{Magnus designed the methodology, conducted the investigation, and was responsible for data curation, formal analysis, validation, and visualization. Magnus wrote the original draft of the manuscript.}

% Address/affiliation
\affiliation[1]{organization={Automation and Control, Aalborg University},
            %addressline={Fredrik Bajers Vej 7C}, 
            city={Aalborg},
%          citysep={}, % Uncomment if no comma needed between city and postcode
            %postcode={9220}, 
            %state={},
            country={Denmark}}

\author[1]{Dorte Hammershøi}%[]
\credit{Dorte Hammershøi co-supervised the study remotely, and contributed i.a. to the discussions of limitations and generalization}
\author[1]{Rafa{\l} Wi{\'s}niewski}%[]
\credit{Rafal supervised the study remotely. Rafal contributed to the design of the methodology, and investigation}

\author[2]{Olga Fink}%[]

% Credit authorship
\credit{Olga hosted the first author as visiting PhD, as well as supervising the research. Olga contributed to the design of the methodology, investigation and provided the resources. All authors reviewed and edited the manuscript and have approved the final version for publication}

% Address/affiliation
\affiliation[2]{organization={Intelligent Maintenance and Operations Systems, EPFL},
            %addressline={1015 Ecublens}, 
            city={Lausanne},
%           citysep={}, % Uncomment if no comma needed between city and postcode
            %postcode={}, 
            %state={},
            country={Switzerland}}

% Footnote text
%\fntext[1]{}

% For a title note without a number/mark
%\nonumnote{}

% Here goes the abstract
\begin{abstract}
Fault detection is critical in industrial systems, enabling early identification of abnormal behaviour and improving safety, reliability, and operational efficiency. Modern industrial systems increasingly use heterogeneous sensing modalities that capture complementary aspects of the underlying physical process. However, existing data-driven anomaly detection methods typically process each modality independently or rely on simple feature-level fusion, failing to exploit cross-modal relationships that characterize normal system behaviour. Their performance also often assumes similar training and deployment distributions. In practice, changing operating conditions, diverse environmental influences, and system degradation induce distribution shifts that can substantially degrade overall detection performance, particularly under unseen operating regimes. Developing multimodal anomaly detection methods that leverage cross-modal information while remaining robust to such shifts therefore remains a major challenge.

In this work, we propose a multimodal anomaly detection framework based on cross-modal reconstruction of heterogeneous time-series sensor data. Rather than modeling each sensing modality independently, the approach learns system dynamics by reconstructing each modality from the others, thereby exploiting complementary information across sensor measurements. This integrates information across sensing channels without requiring explicit temporal alignment or identical sampling rates, while improving robustness to sensor noise, missing measurements, and modality-specific disturbances. To address distribution shifts during real-world deployment, anomalies are identified using cross-modal reconstruction error and an adaptive test-time thresholding mechanism that adjusts to changing operating conditions. Experiments on three industrial case studies demonstrate that the framework consistently achieves strong fault detection performance and substantially improves robustness under out-of-distribution conditions, with the largest gains in the most challenging operating regimes.
\end{abstract}

% Use if graphical abstract is present
%\begin{graphicalabstract}
%\includegraphics{}
%\end{graphicalabstract}

% Research highlights
\begin{highlights}
    \item Multimodal fault detection via cross-modality reconstruction.
    \item No sampling-rate synchronisation required across modalities.
    \item Attention-based fusion enables implicit cross-modal alignment.
    \item Improves robustness under severe out-of-distribution conditions.
\end{highlights}
% Keywords
% Each keyword is seperated by \sep

\maketitle

% Main text
%%%%%%%%%%%%%%%%%%%%%%%%%%%%%%%%%%%%%%%%%%%%%%%%%%%%%%%%%%%%%%%%
\section{Introduction}
%%%%%%%%%%%%%%%%%%%%%%%%%%%%%%%%%%%%%%%%%%%%%%%%%%%%%%%%%%%%%%%%
Reliable fault detection is essential for the safe and efficient operation of engineered systems, as early identification of incipient faults helps prevent failures, reduce unplanned downtime, and improve operational reliability. 
Accurately distinguishing faults from normal operational variability, however, remains challenging because system behavior is inherently dependent on operating conditions. Changes in factors such as speed, load, temperature and environmental conditions can substantially alter the observed sensor measurements, making it difficult to distinguish healthy system behavior from fault-induced deviations across changing operating conditions encountered in practice~\cite{chen2020fault}. Consequently, effective fault detection requires models that distinguish fault-induced deviations from changes in sensor measurements caused by varying operating conditions. 

Model-free approaches have become increasingly prominent for fault detection due to improvements in sensing capabilities and the growing availability of monitoring data~\cite{FINK2020103678, XU2021107530}. However, these methods typically depend on access to training data that adequately cover the range of conditions encountered during operation~\cite{HU2022108063}.This assumption is rarely satisfied in practice. Industrial systems operate under continuously changing operating regimes, environmental conditions, and progressive system degradation, resulting in distribution shifts between training and deployment data that can substantially degrade fault detection performance, particularly under previously unseen operating conditions. Developing fault detection methods that remain reliable under such distribution shifts therefore remains a major challenge.

Existing fault detection approaches attempt to  improve robustness by learning invariant representations or  adapting models when new data becomes available~\cite{michau2021unsupervised,SUN2026112135, michau2019unsupervised}. While these strategies can improve robustness, adaptation-based methods require updating model parameters during deployment, making their performance dependent on the assumptions underlying the adaptation process and the characteristics of the test-time data. 

Beyond improving robustness through adaptation, modern industrial systems are increasingly instrumented with multiple heterogeneous modalities, including vibration, acoustic, electrical, thermal, and process measurements, each capturing complementary aspects of the underlying system dynamics. In principle, jointly exploiting these complementary observations enables a more complete characterization of the underlying system dynamics than any individual modality alone.
However, effectively learning from heterogeneous multimodal data remains challenging due to the differing statistical and structural characteristics of individual modalities, requiring effective fusion strategies~\cite{Zhao2024MultimodalFusion}. In industrial monitoring, these differences often manifest as varying sampling frequencies, noise characteristics, temporal resolutions, and failure mechanisms. Many existing multimodal approaches either process modalities independently or fuse them at the decision level, limiting their ability to learn consistent cross-modal representations that capture the shared physical behavior of the system~\cite{Zhao2024MultimodalFusion}. Furthermore, relatively few existing methods explicitly model how relationships between sensing modalities evolve under changing operating conditions, which can reduce robustness to distribution shifts. Consequently, a multimodal model must not only learn effective cross-modal representations but also ensure that these representations remain invariant across domains while preserving complementary modality-specific information. 

Multi Modal Domain Generalization addresses this problem by learning from multiple source domains and modalities to generalize to unseen target domains without target-domain adaptation. Compared with unimodal domain generalization, MMDG must account for modality-dependent distribution shifts while simultaneously modeling shared and modality-specific information across sensing modalities~\cite{dong2026advances}. Recent methods such as SimMMDG explicitly disentangle modality-shared and modality-specific representations and employ cross-modal translation to improve both domain generalization and robustness to missing modalities~\cite{dong2023simmmdg}. Nevertheless, these approaches have seen limited application in industrial condition monitoring, where heterogeneous sensors are often asynchronous, exhibit different reliability characteristics, and experience domain shifts that vary across modalities.

Consequently, despite the increasing availability of multimodal sensing, existing anomaly detection methods typically address either multimodal representation learning or robustness to distribution shifts in isolation. To the best of our knowledge, no existing framework simultaneously (i) exploits heterogeneous and asynchronous sensor measurements to learn a unified representation of system dynamics and (ii) maintains robustness under severe out-of-distribution operating conditions.

To address this gap, we propose a multimodal framework for time-series fault detection that exploits heterogeneous sensor measurements to learn a shared representation of the underlying system dynamics. The approach is based on cross-modal reconstruction, where each modality is inferred from the others, enabling the model to capture consistent behavior across sensing modalities while integrating complementary information without requiring  synchronized  sampling or explicit temporal alignment. The architecture combines modality-specific encoders with attention-based latent alignment, allowing it to operate directly on raw time series and improving robustness to sensor noise, and missing data. 
Rather than relying on individual sensor measurements, the model learns the underlying cross-modal consistency of normal system behavior, making it substantially less sensitive to changes in operating conditions than conventional reconstruction-based methods. 
During inference, anomalies are detected from the cross-modality reconstruction error using an adaptive test-time threshold that compensates for non-stationary operating conditions and improves robustness to distribution shifts.

\vspace{3mm}

The rest of the paper is organized as follows. 
Section \ref{sec:relatedWork} presents the related work, 
Section \ref{sec:prelim} describes the preliminaries, 
Section \ref{sec:method} describes the proposed method,
Section \ref{sec:experiment} details the experimental setup, 
Section \ref{sec:results} presents the results, 
Section \ref{sec:discussion} presents the discussion, 
and Section \ref{sec:conclusion} concludes the paper.

\section{Related work} \label{sec:relatedWork}

\subsection{Fault Detection under Distribution Shifts}
Prognostics and Health Management (PHM) seeks to improve the reliability and availability of engineered systems by enabling the early detection of faults, accurate diagnosis of their underlying causes, and prediction of the remaining useful life~\cite{FINK2020103678}. In practice, developing data-driven fault detection methods is particularly challenging because representative fault data are inherently scarce. The development of supervised fault detection methods is fundamentally constrained by the limited availability of representative fault data. Industrial assets are designed to operate reliably, causing failures to occur only infrequently, while gradual degradation often spans months or years before culminating in observable faults. As a result, monitoring data are overwhelmingly dominated by healthy operating conditions, leading to severe class imbalance and insufficient fault examples for supervised learning.

As a result, unsupervised anomaly detection has become the predominant paradigm for industrial fault detection~\cite{michau2017deep, liu2018artificial}. Existing methods generally learn a model of healthy system behavior from nominal operating data and identify anomalies as deviations from this learned representation. Depending on how healthy behavior is modeled, these approaches can be broadly categorized into one-class classification, probabilistic modeling, and reconstruction-based methods~\cite{Hsu2023ResidualComparison}. One-class methods, including SVDD~\cite{10.1023/B:MACH.0000008084.60811.49}, One-Class SVM~\cite{10.5555/944790.944808}, and their deep variants such as Deep SVDD~\cite{Ruff2018, frusque2024non}, learn compact decision boundaries that separate healthy observations from potential anomalies. Probabilistic approaches instead estimate the probability distribution of healthy observations and identify anomalous samples based on their likelihood under the learned distribution~\cite{ruff2021unifying}. Reconstruction-based methods, including autoencoders~\cite{10.1145/2689746.2689747}, variational autoencoders~\cite{Iqbal04032023, chao2021implicit}, masked autoencoders~\cite{he2022masked}, and generative adversarial networks~\cite{10.1007/978-3-319-59050-9_12}, assume that models trained exclusively on healthy data accurately reconstruct nominal operating conditions, while faulty observations produce elevated reconstruction errors that serve as anomaly scores.

Despite their methodological differences, these approaches share a common assumption: the distribution of healthy observations encountered during deployment is sufficiently represented by the training data. In practice, however, industrial systems operate under continuously changing loads, environmental conditions, control strategies, and progressive degradation, causing substantial distribution shifts even in the absence of faults~\cite{FINK2020103678}. Consequently, changes in operating conditions may produce reconstruction errors or decision-boundary violations that are indistinguishable from genuine faults, leading to increased false alarms and degraded detection reliability. Developing anomaly detection methods that remain robust under such distribution shifts therefore remains a fundamental challenge.

\subsection{Learning Robust Representations under Distribution Shifts}

To address changing operating conditions, considerable research has focused on learning representations that generalize across domains. Domain adaptation aligns source and target distributions through adversarial learning, discrepancy minimization, or self-training~\cite{pan2009survey, wang2018deep, ganin2016domain}. Within PHM, several approaches have successfully transferred knowledge across operating conditions using adversarial feature alignment and unsupervised adaptation~\cite{michau2021unsupervised}. More recently, test-time and continual adaptation methods have enabled models to update themselves online using only unlabeled target data~\cite{wang2021tent,wang2022continual,SUN2026112135}.

A complementary research direction is domain generalization~\cite{nejjar2022dg, muandet2013domaingeneralizationinvariantfeature}, which aims to learn operating-condition-invariant representations without requiring access to target-domain data during training. Such methods typically exploit multiple source domains, invariant feature learning, or contrastive objectives to improve generalization toward unseen environments.

However, the vast majority of these approaches have been developed for supervised learning tasks, such as fault diagnosis and remaining useful life prediction \cite{nejjar2024domain, nejjar2025uncertainty, wang2019domain, chen:DeepTransferLearning:Review}, where labeled fault data are available in at least the source domain. Their underlying objectives rely on preserving class-discriminative information across domains, making them difficult to transfer directly to unsupervised anomaly detection, where only healthy operating data are available during training. While recent work has begun to investigate domain adaptation for unsupervised fault detection~\cite{michau2021unsupervised, SUN2026112135}, relatively few methods explicitly address this setting.

Moreover, both domain adaptation and domain generalization remain fundamentally constrained by the variability available during training or adaptation. Domain adaptation assumes that representative target-domain observations become available during deployment, whereas domain generalization relies on sufficiently diverse source domains to learn operating-condition-invariant representations. In practical industrial systems, however, operating conditions evolve continuously throughout the asset lifetime, often giving rise to previously unseen operating regimes that cannot be anticipated during training or adequately captured during online adaptation. Consequently, developing unsupervised fault detection methods that remain robust under severe operating-condition distribution shifts remains an open challenge.

\subsection{Multimodal Representation Learning for Fault Detection}
Modern industrial systems are increasingly instrumented with heterogeneous sensing modalities, including vibration, acoustic, electrical, thermal, and process measurements. Since these sensors capture complementary aspects of the underlying physical processes, multimodal learning has emerged as a promising direction for improving fault detection performance~\cite{Zhao2024MultimodalFusion,10.1145/3711896.3736567}.

Existing multimodal approaches typically combine information through early fusion, late fusion, or shared latent representations~\cite{Zhao2024MultimodalFusion, Wu2024MLMM}. More recent architectures introduce modality-specific encoders before fusion, enabling each sensing modality to preserve its characteristic temporal and spectral properties while learning joint representations~\cite{Zhao2024MultimodalFusion,Kibrete2024Review}. Such designs have demonstrated improved robustness to missing modalities, sensor failures, and heterogeneous noise characteristics.

However, existing multimodal fault detection methods have primarily focused on improving prediction accuracy under nominal operating conditions rather than robustness under distribution shifts. Most approaches implicitly assume that the relationships between sensing modalities learned during training remain valid during deployment. When operating conditions change, however, these cross-modal relationships may themselves evolve, reducing the effectiveness of learned multimodal representations. Consequently, the potential of multimodal sensing for improving robustness under changing operating conditions remains largely unexplored.

A further challenge arises from the heterogeneous temporal characteristics of industrial sensing systems. Different sensors frequently operate at different sampling rates and exhibit asynchronous observations~\cite{LIN2019335}. Existing multirate fusion methods generally transform asynchronous observations into synchronized representations before fusion~\cite{LIN2019335}. While effective, explicit synchronization introduces interpolation errors, communication delays, and additional uncertainty that may degrade downstream anomaly detection performance. Developing multimodal learning frameworks that directly accommodate asynchronous sensing therefore remains an important open research problem.

\subsection{Cross-modal Self-supervised Learning}

Self-supervised learning has emerged as a powerful paradigm for learning multimodal representations without requiring manual annotations by exploiting the intrinsic relationships between complementary sensing modalities. Rather than relying on semantic labels, these methods leverage cross-modal consistency as a supervisory signal, encouraging representations to capture the latent physical processes that simultaneously give rise to multiple observations. As a result, multimodal self-supervised learning has demonstrated remarkable success across computer vision, robotics, and multimodal perception, where complementary sensing modalities provide multiple views of the same underlying scene or physical process.

Existing multimodal self-supervised approaches can broadly be divided into joint reconstruction and cross-modal reconstruction. Joint reconstruction methods learn a shared latent representation from multiple sensing modalities and reconstruct all modalities simultaneously. Such formulations have been applied to robotic anomaly detection using multimodal variational autoencoders that jointly model visual, force, audio, and kinematic observations~\cite{8279425, 10.1109/ICRA.2019.8793485}. Although these methods implicitly capture cross-modal relationships through a shared latent space, their reconstruction objective is dominated by recovering individual modalities rather than explicitly modeling the consistency between them. Consequently, they provide only limited supervision for learning modality-invariant representations that reflect the shared system dynamics.

Cross-modal reconstruction addresses this limitation by reconstructing one modality exclusively from complementary observations. This objective forces the latent representation to encode information that is common across sensing modalities while suppressing sensor-specific noise and artifacts, making it particularly attractive for learning representations of the underlying physical system rather than individual sensor characteristics. Cross-modal reconstruction has recently shown promising results in computer vision and multimodal perception. For example, CMDR-IAD employs bidirectional reconstruction between RGB imagery and 3D geometry for industrial anomaly detection, demonstrating improved robustness to noisy observations and missing modalities~\cite{Daci2026CMDRIAD}. Similarly, SimMMDG explicitly decouples features into modality-shared and modality-specific components while employing cross-modal translation to enforce consistency and maintain robustness against missing modalities under distribution shifts~\cite{dong2023simmmdg}.

Despite its promise, cross-modal reconstruction has received comparatively little attention for industrial multivariate time-series fault detection. Existing approaches have primarily been developed for visual modalities and generally assume synchronized observations acquired under relatively stationary conditions. In contrast, industrial monitoring systems consist of heterogeneous sensors operating at different sampling rates, exhibiting modality-specific noise characteristics, and continuously evolving operating conditions that fundamentally alter the relationships between sensing modalities. Consequently, current cross-modal learning frameworks neither explicitly model asynchronous multimodal time-series nor learn representations that remain robust under severe operating-condition distribution shifts. Developing self-supervised representations that exploit cross-modal consistency while accommodating heterogeneous sensing and changing operating conditions therefore remains an important open research challenge.

%%%%%%%%%%%%%%%%%%%%%%%%%%%%%%%%%%%%%%%%%%%%%%%%%%%%%%%%%%%%%%%%
\section{Preliminaries} \label{sec:prelim}
%%%%%%%%%%%%%%%%%%%%%%%%%%%%%%%%%%%%%%%%%%%%%%%%%%%%%%%%%%%%%%%%
In the following, we introduce the modeling framework and assumptions underlying the proposed method. These assumptions are adopted as methodological design choices that allow us to isolate and analyze the effect of operating conditions on anomaly detection performance.

We consider a discrete-time dynamical system operating under varying operating conditions. The system evolves according to

\begin{equation}
\begin{cases}
x_{k+1} = f_{\rho}(x_k, w_k), \\
y_k = h_{\rho}(x_k) + v_k,
\end{cases}
\end{equation}
where $x_k \in \mathbb{R}^n$ denotes latent system state, $y_k \in \mathbb{R}^r$ the measured observations and $\rho_k \in \mathcal{P}$ the operating condition at time step $k$, with $\mathcal{P}$ denoting the set of discrete operating conditions considered in this work. The terms $w_k$ and $v_k$ represent process and measurement noise, respectively. %The terms $w_k$ and $v_k$ represent process and measurement noise, respectively.

The operating condition influences both the system evolution and the observation process, resulting in changes to the distribution of sensor measurements even when the system remains healthy. Consequently, observations acquired under different operating conditions may exhibit substantially different statistical characteristics despite corresponding to the same underlying health state. This variability constitutes the primary source of distribution shift considered in this work.

We further assume that all sensing modalities are generated from the same latent system state $x_k$, while each modality provides only a partial and modality-specific observation of this  state through. Consequently, no individual sensing modality is assumed to fully characterize the system dynamics. Instead, the complementary information provided by multiple sensing modalities motivates learning a shared latent representation that captures the underlying system behavior while remaining robust to changes in operating conditions.

%%%%%%%%%%%%%%%%%%%%%%%%%%%%%%%%%%%%%%%%%%%%%%%%%%%%%%%%%%%%%%%%
\subsection{Operational Parameters}
\begin{figure}[pos=h!]
\centering
\resizebox{\linewidth}{!}{%
    \begin{tikzpicture}[line width=0.8pt]
        
        % First few steps
        \draw (0,0) -- (1.5,0);                     % step 1
        \draw (1.5,0) -- (1.5,1) -- (4,1);          % step 2
        \draw (4,1) -- (4,0.5) -- (6,0.5);          % step 3
        \draw (6,0.5) -- (6,0) -- (7,0);            % step 4 start
        
        % Ellipsis for continuation
        \node at (7.55,0) {$\ldots$};
        \node at (8.05,0) {$\ldots$};
        %\node at (7.8,1) {$\vdots$};
        
        % Final step (k-th)
        \draw (9,0) -- (9,0.7) -- (11,0.7);   
        \node[circle,draw,inner sep=1pt] at (10,0.9) {$k$};
        
        % Circle labels for early states
        \node[circle,draw,inner sep=1pt] at (0.75,0.2) {1};
        \node[circle,draw,inner sep=1pt] at (2.75,1.2) {2};
        \node[circle,draw,inner sep=1pt] at (5,0.7) {3};
        
    \end{tikzpicture}%
}
\caption{Illustration of the operating-condition switching sequence in a parameter-varying system with a finite set operational parameters}
\label{fig:Scheduling}
\end{figure}
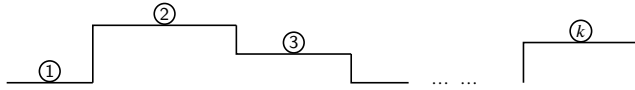

%%%%%%%%%%%%%%%%%%%%%%%%%%%%%%%%%%%%%%%%%%%%%%%%%%%%%%%%%%%%%%%%
We consider systems operating under multiple operating conditions, such as different loads, speeds, temperatures, or controller settings. The operating condition is represented by a parameter $\rho$, which is assumed to take values from a finite set of discrete operating modes and to be known at each time step. Changes between operating conditions occur instantaneously over time, as illustrated in Figure~\ref{fig:Scheduling}. This formulation enables us to explicitly separate changes in the observed measurements caused by varying operating conditions from those resulting from changes in the underlying health state.

Lastly, we consider the inference scenario in which the system encounters operating conditions that were not observed during training. To study generalization under such distribution shifts, we adopt the following assumption: if the system state at the preceding time step is classified as fault-free, then the system remains fault-free immediately after the change in operating condition. This assumption is introduced to enable the study of fault-free generalization across unseen operating conditions without requiring explicit modeling of conditional or adaptive operating dynamics.

Under this assumption, the objective of the proposed method is to learn a representation of system health that generalizes across operating conditions, while allowing the resulting anomaly scores to remain dependent on the operating condition.

%%%%%%%%%%%%%%%%%%%%%%%%%%%%%%%%%%%%%%%%%%%%%%%%%%%%%%%%%%%%%%%%
\subsection{Reconstruction-based fault detection}\label{sc:reconstructionbasedFaultDetection}
%%%%%%%%%%%%%%%%%%%%%%%%%%%%%%%%%%%%%%%%%%%%%%%%%%%%%%%%%%%%%%%%
We employ a multimodal, reconstruction‑based framework for unsupervised anomaly detection, as illustrated in Figure~\ref{fig:architecture}. The underlying assumption is that a model trained exclusively on normal (healthy) data learns the intrinsic relationships between the available sensing modalities and therefore captures the characteristics of nominal system operation. During deployment, measurements that cannot be accurately reconstructed from these learned cross-modal relationships indicate a deviation from normal behavior and are consequently interpreted as potential anomalies.

Unlike conventional autoencoder-based approaches that reconstruct each modality from itself, we formulate anomaly detection as a cross-modal reconstruction problem. Specifically, one sensing modality is reconstructed from complementary modalities. This formulation encourages the model to exploit the physical and statistical dependencies between heterogeneous sensor streams rather than relying on modality-specific patterns that may also reproduce anomalous observations. As a result, anomalies that disrupt the consistency between modalities produce larger reconstruction errors, improving the discriminative power of the learned representation.

Let $\mathbf{x}^{(1)}, \ldots, \mathbf{x}^{(M)}$ denote the $M$ input modalities, where $\mathbf{x}^{(m)} \in \mathbb{R}^{L \times d_m}$ represents a windowed time series of length $L$ for modality $m$, and $d_m$ is the dimensionality of the modality. During training, the model learns a mapping

\begin{equation}
    \hat{\mathbf{x}}^{(m)} =
    f_\theta\!\left(
    \mathbf{x}^{(1)}, \ldots, \mathbf{x}^{(m-1)},
    \mathbf{x}^{(m+1)}, \ldots, \mathbf{x}^{(M)}
    \right).
\end{equation}

where $f_\theta(\cdot)$ denotes the multimodal reconstruction network parameterized by $\theta$, which learns to approximate the mapping from the remaining sensing modalities to the target modality. The output $\hat{\mathbf{x}}^{(m)} \in \mathbb{R}^{L \times d_m}$  denotes the reconstructed windowed time series of modality $m$. The network parameters are optimized using only healthy operating data by minimizing a reconstruction loss
\begin{equation}
    \mathcal{L}_{\mathrm{rec}}
    =
    \frac{1}{N}\sum_{n=1}^{N}
    \ell\!\left(
    \mathbf{x}^{(m)},
    \hat{\mathbf{x}}^{(m)}
    \right),
\end{equation}
where $\ell(\cdot,\cdot)$ denotes the reconstruction error. During inference, the reconstruction error serves as an anomaly score, with larger errors indicating a greater likelihood that the observed measurements deviate from the learned normal operating behavior.

%%%%%%%%%%%%%%%%%%%%%%%%%%%%%%%%%%%%%%%%%%%%%%%%%%%%%%%%%%%%%%%%
\section{Proposed method}
\label{sec:method} 
\begin{figure*}[t]
    \centering
    \includegraphics[width=0.85\textwidth,trim={0cm 6.5cm 0cm 0cm},clip]{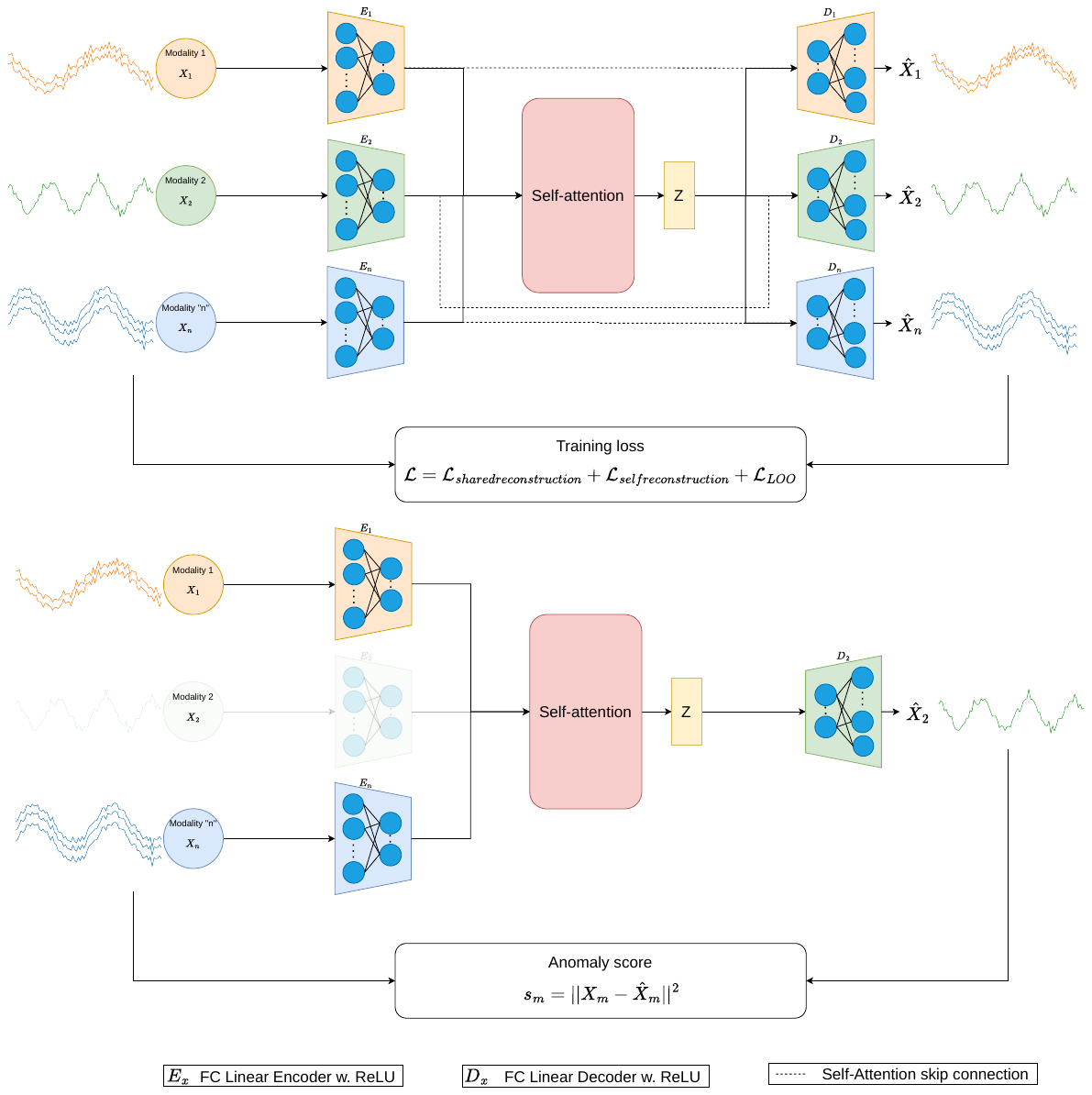}
    \caption{Training and inference architecture with modality-specific encoders and decoders, an attention-based fusion module, and an optional self-reconstruction path, which bypasses the fusion layer.}
    \label{fig:architecture}
\end{figure*}

%%%%%%%%%%%%%%%%%%%%%%%%%%%%%%%%%%%%%%%%%%%%%%%%%%%%%%%%%%%%%%%%
This section presents the proposed method, detailing the model architecture, training objectives, and inference procedure, while the online threshold adaptation is introduced in Section~\ref{sc:inferenceAndUpdate}. Building on the cross-modal reconstruction framework introduced in Section~\ref{sc:reconstructionbasedFaultDetection}, the proposed model implements the reconstruction function $f_\theta$ using a multi-encoder, multi-decoder autoencoder architecture. The overall training architecture is illustrated in Figure~\ref{fig:architecture}, and the corresponding objective functions are described in Section~\ref{sc:trainingObjectives}.

The model is trained exclusively on healthy samples from the source domain $X_s$. Each encoder processes one input modality, while the decoders reconstruct the target modality from the latent representations of the remaining modalities, yielding reconstructed outputs $\hat{X}_s$.

During inference, the encoder corresponding to the target modality is removed so that the omitted modality is reconstructed solely from the remaining sensor streams, as illustrated in Figure~\ref{fig:architecture}. Although any modality can be selected as the reconstruction target, the achievable reconstruction accuracy depends on the predictive information contained in the available modalities.

%%%%%%%%%%%%%%%%%%%%%%%%%%%%%%%%%%%%%%%%%%%%%%%%%%%%%%%%%%%%%%%%
\subsection{Definitions and training} \label{sc:trainingObjectives}
%%%%%%%%%%%%%%%%%%%%%%%%%%%%%%%%%%%%%%%%%%%%%%%%%%%%%%%%%%%%%%%%
Let $\mathbf{x}^{(1)}, \ldots, \mathbf{x}^{(M)}$ denote the $M$ input modalities, where $\mathbf{x}^{(m)} \in \mathbb{R}^{L \times d_m}$ represents a windowed time series of modality $m$. Each modality is encoded into a modality-specific latent representation
\begin{equation*}
    \mathbf{z}^{(m)} = E_m\!\left(\mathbf{x}^{(m)}\right).
\end{equation*}

A shared latent representation is obtained using a self-attention fusion module
\begin{equation*}
    \mathbf{z}_{\mathrm{shared}}
    =
    \mathcal{A}\!\left(
    \mathbf{z}^{(1)},\ldots,\mathbf{z}^{(M)}
    \right).
\end{equation*}  

where $\mathcal{A}(\cdot)$ denotes a multi-head self-attention fusion operator. Given the modality-specific latent representations
\[
\mathbf H=
[\mathbf z^{(1)}+\mathbf e^{(1)},\ldots,\mathbf z^{(M)}+\mathbf e^{(M)}],
\]
the fused latent representation is computed as
\begin{equation}
\mathbf z_{\mathrm{shared}}
=
\operatorname{Mean}
\left(
\operatorname{LayerNorm}
\left(
\mathbf H+\operatorname{MHA}(\mathbf H)
\right)
\right),
\end{equation}
where $\mathbf e^{(m)}$ denotes a learnable modality embedding and $\operatorname{Mean}(\cdot)$ averages over the modality tokens.

Since the attention module preserves the latent dimensionality, all decoders receive inputs of identical size. Consequently, any modality can be removed during training or inference without requiring architectural modifications. This property further enables a leave-one-out reconstruction strategy. Each decoder reconstructs its target modality from a shared latent representation, computed from the remaining available modalities.

\begin{equation*}
    \hat{\mathbf{x}}^{(m)}
    =
    D_m\!\left(
    \mathcal{A}\!\left(
    \mathbf{z}^{(1)}, \ldots,
    \mathbf{z}^{(m-1)},
    \mathbf{z}^{(m+1)}, \ldots,
    \mathbf{z}^{(M)}
    \right)
    \right).
\end{equation*}

where $z$ may denote a modality-specific latent representation, the shared latent representation, or a leave-one-out shared latent representation. It is important to note, that if $M = 1$, the attention matrix is necessary $[1]$, and attention performs no cross-modal selection. The fusion block reduces to a learned transformation of the single remaining modality followed by residual normalization.

Training optimizes three objectives - self-reconstruction loss, shared-latent reconstruction loss and leave-one-out cross-modal loss: 
(i) Self-reconstruction is employed to preserve modality-specific structure via a bypass path, mitigating the risk that fusion attenuates modality-unique cues,
(ii) Shared-latent reconstruction to ensure that fused latents carry enough information to reconstruct each modality, and 
(iii) Cross modality reconstruction (leave-one-out reconstruction) which operationalizes the “any subset may be omitted at inference” requirement, while also enforcing a unified latent space. 

\paragraph{Self-reconstruction loss}
Each modality is reconstructed from its corresponding latent representation:
\begin{equation}
    \mathcal{L}_{\text{self}} = \sum_{m=1}^{n} \lVert x_m - D_m(E_m(x_m)) \rVert^2.
\end{equation}

\paragraph{Shared-latent reconstruction loss}
The shared latent representation is used to reconstruct all modalities:
\begin{equation}
    \mathcal{L}_{\text{shared}} = \sum_{m=1}^{n} \lVert x_m - D_m(z_{\text{shared}}) \rVert^2.
\end{equation}

\paragraph{Leave-one-out cross-modal loss}
For each target modality, the shared latent representation is recomputed by applying the self-attention fusion operator to the latent representations of all remaining modalities,
\begin{equation}
    \mathbf{z}_{\mathrm{shared}}^{(m)}
    =
    \mathcal{A}\!\left(
    \mathbf{z}^{(1)}, \ldots,
    \mathbf{z}^{(m-1)},
    \mathbf{z}^{(m+1)}, \ldots,
    \mathbf{z}^{(M)}
    \right).
\end{equation}

The leave-one-out reconstruction loss is defined as
\begin{equation}
    \mathcal{L}_{\mathrm{loo}}
    =
    \sum_{m=1}^{M}
    \left\|
    \mathbf{x}^{(m)}
    -
    D_m\!\left(\mathbf{z}_{\mathrm{shared}}^{(m)}\right)
    \right\|^2
\end{equation}

\paragraph{Total loss}
The overall objective is defined as
\begin{equation}
    \mathcal{L} = \lambda_1 \mathcal{L}_{\text{self}}
                + \lambda_2 \mathcal{L}_{\text{shared}}
                + \lambda_3 \mathcal{L}_{\text{loo}}.
\end{equation}

\paragraph{Anomaly score}
During inference, one modality is omitted and reconstructed from the remaining modalities:
\begin{equation}
    \hat{x}_m = D_m\left(z_{\text{shared}}^{(-m)}\right).
\end{equation}

The anomaly score for modality $m$ is defined as the reconstruction error:
\begin{equation}
    s_m = \lVert x_m - \hat{x}_m \rVert^2.
\end{equation}

%%%%%%%%%%%%%%%%%%%%%%%%%%%%%%%%%%%%%%%%%%%%%%%%%%%%%%%%%%%%%%%%
\subsection{Inference with Adaptive Test-Time threshold}\label{sc:inferenceAndUpdate}
%%%%%%%%%%%%%%%%%%%%%%%%%%%%%%%%%%%%%%%%%%%%%%%%%%%%%%%%%%%%%%%%
As we will demonstrate later, the AUC scores reported in Section~\ref{sc:AUCResults:KAIST}, Section~\ref{sec:AMPERE_results} and Section~\ref{sc:AUCResults:IMAD} show that detection performance remains high when evaluated separately for each operating condition, even if the corresponding condition is absent during training. In contrast, aggregating samples across different operating conditions can degrade performance due to distributional shifts between modes. These shifts primarily affect the calibration of the anomaly scores rather than their class separability.

\begin{figure}[pos = h]
    \centering
    \includegraphics[width=\linewidth]{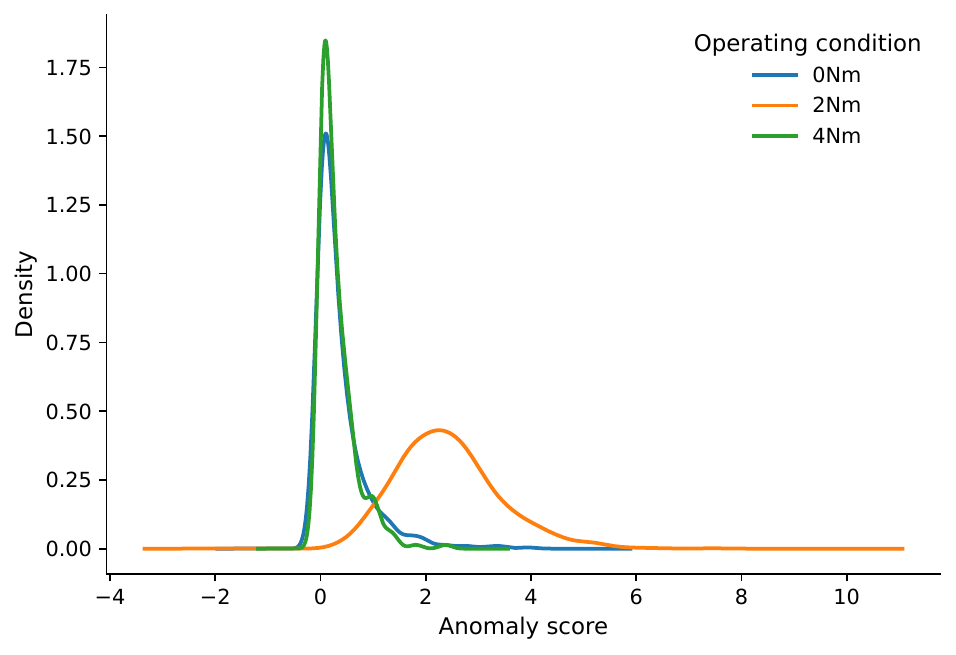}
    \caption{Anomaly score distribution for healthy test data from the KAIST dataset, trained without $2\,\mathrm{Nm}$ data.}
    \label{fig:kdePlotOfNormalDistribution}
\end{figure}

This effect is illustrated in Figure~\ref{fig:kdePlotOfNormalDistribution}, which shows the distribution of anomaly scores for healthy test data in an experiment where the $2\,\mathrm{Nm}$ operating condition was not included during training. The score distributions exhibit systematic shifts across operating conditions, making the use of a single global decision threshold impractical.

This suggests that a key challenge in the considered setting is that the reconstruction score depends not only on the health state of the system, but also on the operating condition~$\rho$. Therefor, we propose using an Adaptive Test-Time threshold. The anomaly score is defined as
\begin{equation}
    s_k = \ell\!\left(\mathbf{y}_k,\, f_{\theta}(\mathbf{y}_k)\right),
\end{equation}

where $f_{\theta}(\cdot)$ is the reconstruction model and $\ell(\cdot,\cdot)$ is the scoring function.

For each operating condition $\rho$, the method maintains a condition-specific state consisting of the local mean, variance, and standard deviation of healthy scores,
\begin{equation}
    \bigl(\hat{\mu}^{(\rho)}_k,\hat{v}^{(\rho)}_k,\hat{\sigma}^{(\rho)}_k\bigr),
    \qquad
    \hat{\sigma}^{(\rho)}_k = \max\!\bigl(\sqrt{\hat{v}^{(\rho)}_k},\,\sigma_{\min}\bigr),
\end{equation}
where $\sigma_{\min}>0$ is a lower bound that prevents degenerate dispersion estimates. Let $\rho_k$ denote the operating condition associated with window $\mathbf{y}_k$.

When a condition $\rho$ is encountered for the first time during deployment, its state is initialized from healthy validation statistics. If condition-specific validation statistics are available, they are used; otherwise, a global healthy baseline is adopted. Specifically,
\begin{align*}
    \hat{\mu}^{(\rho)}_0 &= \mu^{(\rho)}_{\mathrm{init}},
    \\
    \hat{v}^{(\rho)}_0& = \max\!\bigl(v^{(\rho)}_{\mathrm{init}},\,\sigma_{\min}^2\bigr),
    \\
    \hat{\sigma}^{(\rho)}_0 &= \max\!\bigl(\sigma^{(\rho)}_{\mathrm{init}},\,\sigma_{\min}\bigr).
\end{align*}
For operating conditions regarded as out-of-distribution, the same global healthy initialization is used, but the alarm multiplier may differ from that used for in-distribution conditions.

At deployment step $k$, let $\rho_k$ be the current operating condition. Before any adaptation is performed, the detector computes the condition-dependent alarm threshold
\begin{equation}
    \tau_k^- = \hat{\mu}^{(\rho_k)}_{k-1}
    + \kappa_{\mathrm{alarm}}^{(\rho_k)} \hat{\sigma}^{(\rho_k)}_{k-1},
\end{equation}

and declares an anomaly according to
\begin{equation}
    \hat{a}_k = \mathbb{I}\!\left\{ s_k > \tau_k^- \right\}.
\end{equation}
Hence, the detection decision is always based on the pre-update state and is therefore not influenced by the current sample.

To prevent abnormal samples from contaminating the running statistics, adaptation is controlled by a second threshold,
\begin{equation}
    \tau_k^{\mathrm{upd}} =
    \hat{\mu}^{(\rho_k)}_{k-1}
    + \kappa_{\mathrm{upd}} \hat{\sigma}^{(\rho_k)}_{k-1},
\end{equation}

with $\kappa_{\mathrm{upd}} \leq \kappa_{\mathrm{alarm}}^{(\rho)}$. The corresponding update indicator is
\begin{equation}
    g_k = \mathbb{I}\!\left\{ s_k \leq \tau_k^{\mathrm{upd}} \right\}.
\end{equation}

Only samples that remain below the update gate are treated as sufficiently likely to be healthy and are therefore used for adaptation.

If $g_k = 1$, the local mean and variance are updated by exponential smoothing:
\begin{align}
    \hat{\mu}^{(\rho_k)}_k
    &= (1-\alpha)\hat{\mu}^{(\rho_k)}_{k-1} + \alpha s_k, \\
    \hat{v}^{(\rho_k)}_k
    &= (1-\alpha)\hat{v}^{(\rho_k)}_{k-1}
    + \alpha \bigl(s_k - \hat{\mu}^{(\rho_k)}_{k-1}\bigr)^2,
\end{align}

where $\alpha \in (0,1]$ is the adaptation rate. The corresponding standard deviation estimate is then
\begin{equation}
    \hat{\sigma}^{(\rho_k)}_k
    = \max\!\bigl(\sqrt{\hat{v}^{(\rho_k)}_k},\,\sigma_{\min}\bigr).
\end{equation}

If $g_k = 0$, no adaptation is performed and the state remains unchanged:
\begin{equation}
    \hat{\mu}^{(\rho_k)}_k = \hat{\mu}^{(\rho_k)}_{k-1},
    \qquad
    \hat{v}^{(\rho_k)}_k = \hat{v}^{(\rho_k)}_{k-1},
    \qquad
    \hat{\sigma}^{(\rho_k)}_k = \hat{\sigma}^{(\rho_k)}_{k-1}.
\end{equation}

The updated threshold to be used for subsequent windows under the same operating condition is therefore
\begin{equation}
    \tau_k^+ = \hat{\mu}^{(\rho_k)}_k
    + \kappa_{\mathrm{alarm}}^{(\rho_k)} \hat{\sigma}^{(\rho_k)}_k.
\end{equation}

Overall, the proposed method implements a condition-wise, gated exponentially weighted threshold. The threshold tracks gradual score drift associated with changing operating conditions, whereas large score excursions are excluded from the adaptation process and remain available for anomaly detection. By decoupling the alarm threshold parameter $\kappa_{\mathrm{alarm}}^{(\rho)}$ from the update-gate parameter $\kappa_{\mathrm{upd}}$, the method provides explicit control over the trade-off between detection sensitivity and adaptation robustness. This property is particularly relevant in deployment settings characterized by evolving operating conditions and limited labeled fault data.

%%%%%%%%%%%%%%%%%%%%%%%%%%%%%%%%%%%%%%%%%%%%%%%%%%%%%%%%%%%%%%%%
\section{Experimental Setup} \label{sec:experiment}
%%%%%%%%%%%%%%%%%%%%%%%%%%%%%%%%%%%%%%%%%%%%%%%%%%%%%%%%%%%%%%%%
\subsection{Datasets}

We evaluate the proposed framework on three publicly available multimodal datasets that cover different industrial systems, sensing modalities, fault types, and operating conditions:

(i) the AMPERE dataset~\cite{soualhi2023ampere},
(ii) the Vibration, Acoustic, Temperature, and Motor Current Dataset of Rotating Machinery under Varying Operating Conditions for Fault Diagnosis~\cite{JUNG2023109049}, referred to as the KAIST dataset, and
(iii) the IMAD-DS dataset~\cite{augusti_2024_12665499}.

The selected datasets cover diverse industrial systems, sensing modalities, fault types, and operating conditions, enabling a comprehensive evaluation of the proposed framework under both nominal and distribution-shift scenarios.
%%%%%%%%%%%%%%%%%%%%%%%%%%%%%%%%%%%%%%%%%%%%%%%%%%%%%%%%%%%%%%%%
\subsubsection{AMPERE}
The AMPERE dataset is acquired from a laboratory test bench designed for comprehensive monitoring of rotor- and stator-related faults in a three-phase squirrel-cage induction motor~\cite{soualhi2023ampere}. In this work, we focus exclusively on rotor-related faults.

The experimental setup consists of a 5.5\,kW induction motor driven by a three-phase inverter and coupled to an electromagnetic brake capable of dissipating up to 5\,kW and generating a maximum torque of 100\,Nm, enabling controlled variation of load conditions. Measurements are collected using a heterogeneous sensor suite comprising three-phase voltage and current sensors, three orthogonally mounted accelerometers, and a rotational speed encoder.

The dataset contains 25 experiments covering five health conditions, including healthy operation, one-, three-, and four-bar rotor breakages, together with bearing faults, evaluated under five distinct load levels. Unless otherwise stated, experiments initially focus on the least severe fault (one broken rotor bar), while the final evaluation considers all fault types.

%%%%%%%%%%%%%%%%%%%%%%%%%%%%%%%%%%%%%%%%%%%%%%%%%%%%%%%%%%%%%%%%
\subsubsection{KAIST}

The KAIST rotating machinery dataset~\cite{JUNG2023109049} consists of multimodal measurements acquired from a laboratory rotating machinery test bench under systematically varied operating conditions.

Measurements include vibration, acoustic, temperature, motor current, and rotational speed signals. In this work, vibration, temperature, and motor current measurements are utilized. Rotational speed is excluded because it remains constant across experiments, while acoustic measurements are omitted due to incomplete availability across operating conditions.

The dataset contains both healthy and faulty operating conditions, including bearing defects with varying severity, shaft misalignment, and rotor imbalance, evaluated under three load conditions (0, 2, and 4\,Nm).

%%%%%%%%%%%%%%%%%%%%%%%%%%%%%%%%%%%%%%%%%%%%%%%%%%%%%%%%%%%%%%%%
\subsubsection{IMAD-DS}

The IMAD-DS dataset~\cite{augusti_2024_12665499} evaluates multimodal anomaly detection on a brushless motor operating under varying rotational speeds and acoustic environments.

Measurements consist of synchronized microphone, accelerometer, and gyroscope signals. Domain shifts are induced through changes in rotational speed together with varying levels of background acoustic noise, while anomalies correspond to controlled mechanical perturbations.

Following the unsupervised setting, training is performed exclusively using healthy source-domain data, whereas evaluation includes both healthy and anomalous observations from source and target domains.

%%%%%%%%%%%%%%%%%%%%%%%%%%%%%%%%%%%%%%%%%%%%%%%%%%%%%%%%%%%%%%%%

\subsection{Evaluation Protocol}

Unless stated otherwise, all experiments follow the same evaluation protocol. For the AMPERE and KAIST datasets, healthy operating data are divided into 60\%, 20\%, and 20\% training, validation, and test subsets, respectively, while faulty observations are reserved exclusively for evaluation.

Time-series measurements are segmented into sliding windows of 100\,ms, after which the resulting samples are randomly shuffled. This assumes that the extracted windows can be treated as approximately independent observations originating from the same continuous recording.

To evaluate robustness under changing operating conditions, we consider three evaluation settings:

\begin{itemize}
    \item \textbf{In-distribution (ID):} all operating conditions are available during training.
    \item \textbf{Moderate distribution shift (OOD-M):} intermediate operating conditions are excluded during training and encountered only during testing.
    \item \textbf{Severe distribution shift (OOD-S):} the highest operating conditions are excluded during training and encountered only during testing.
\end{itemize}

For the KAIST and AMPERE datasets, excluded operating conditions are removed entirely from the training and validation sets and are reserved exclusively for evaluation.

For the IMAD-DS dataset, we adopt the experimental protocol proposed in~\cite{augusti_2024_12665499}. However, unlike the original work, target-domain data are strictly excluded during training to maintain a consistent evaluation protocol across all datasets.

%%%%%%%%%%%%%%%%%%%%%%%%%%%%%%%%%%%%%%%%%%%%%%%%%%%%%%%%%%%%%%%%
\subsection{Baselines}

To the best of our knowledge, no existing method directly addresses multimodal cross-modal reconstruction for unsupervised fault detection under operating-condition distribution shifts. We therefore compare the proposed framework against a reconstruction-based baseline employing early sensor fusion through feature concatenation. In addition, we report results obtained using each sensing modality individually to quantify the benefit of multimodal representation learning.

%%%%%%%%%%%%%%%%%%%%%%%%%%%%%%%%%%%%%%%%%%%%%%%%%%%%%%%%%%%%%%%%
\subsection{Implementation Details}

Model hyperparameters are optimized using Optuna~\cite{akiba2019optunanextgenerationhyperparameteroptimization}, where the validation objective is defined as the leave-one-out cross-modal reconstruction performance. The parameters of the adaptive test-time threshold are selected using grid search using the validation F1 score as the optimization objective. The complete set of hyperparameters used in all experiments is provided in Appendix~\ref{sc:app:Hyperparameters}.
%%%%%%%%%%%%%%%%%%%%%%%%%%%%%%%%%%%%%%%%%%%%%%%%%%%%%%%%%%%%%%%%
\section{Results}\label{sec:results}
%%%%%%%%%%%%%%%%%%%%%%%%%%%%%%%%%%%%%%%%%%%%%%%%%%%%%%%%%%%%%%%%
In the following, we present the results obtained on the three datasets described previously.

As the primary objective of this work is to assess robustness under domain shift, the most informative comparisons are those conducted under severe out-of-distribution conditions. These are extensively evaluated on the AMPERE dataset (Section~\ref{sec:AMPERE_results}), where varying levels of distribution shift can be systematically controlled. Given that the observed behavior is consistent in these settings, we do not repeat the full set of baseline comparisons for the KAIST and IMAD-DS datasets in order to avoid redundancy.

AUC scores are reported both per operating condition and overall. The per operating condition AUC values are computed independently for each operating condition, whereas the overall AUC is computed from the pooled predictions across all operating condition. Consequently, the overall AUC is not the average of the per operating condition values, but reflects the model's ability to discriminate normal and faulty samples across the full range of operating conditions.

%%%%%%%%%%%%%%%%%%%%%%%%%%%%%%%%%%%%%%%%%%%%%%%%%%%%%%%%%%%%%%%%
\subsection{Results obtained on AMPERE}\label{sec:AMPERE_results}
%%%%%%%%%%%%%%%%%%%%%%%%%%%%%%%%%%%%%%%%%%%%%%%%%%%%%%%%%%%%%%%%
In Table~\ref{tab:AMPEREAUC} we report AUC scores, for the AMPERE dataset, obtained with: 
(i) the proposed method, 
(ii) an early stage fusion, and
(iii) the best single modality score obtained. In Table~\ref{tab:ampere_overall_summary} we report Accuracy, Precision, Recall and F1-score using the proposed adaptive threshold. 

When a row is denoted $\text{OOD}_m$, we excluded $25\%$ and $75\%$ load condition from the training. When denoted $OOD_{t}$ we excluded $50\%, 75\%$ and $100\%$ load from training. 

In addition to these AUC scores, we report an ablation study in Table~\ref{tab:ablation_ampere_combined}, that shows the contribution of each term in the loss-function - do note though, this is before optimizing with OPTUNA. 

As demonstrated by the reported AUC scores, the proposed method exhibits improved robustness with increasing degrees of out-of-distribution conditions, particularly under severe domain shifts ($\text{OOD}_t$). In these regimes, the multimodal formulation maintains substantially higher detection performance compared to both single-modality and early-fusion approaches, whereas baseline methods degrade significantly. In less challenging settings ($\text{OOD}_m$), performance is generally comparable across methods.

\begin{table}[pos=h]
\centering
\caption{AUC scores obtained with the proposed architecture on the AMPERE dataset, across operational parameters. Results obtained with the fault type "One broken rotor bar". $\text{OOD}_m$ meaning, 25\% and 75\% load is not present during training, $\text{OOD}_t$ meaning 50\%, 75\% and 100\% load is not present during training.}
\label{tab:AMPEREAUC}
\resizebox{\columnwidth}{!}{%
\begin{tabular}{l l ccc}
\toprule
Setting & Load & Proposed method  & Unimodal AE AUC & Early fusion\\
\midrule
\multirow{6}{*}{ID}
    & 0 \%   & 1.0000 & 0.9987 & 0.9980 \\
    & 25 \%  & 0.9990 & 0.9833 & 0.9826 \\
    & 50 \%  & 0.9959 & 0.9979 & 0.9998 \\
    & 75 \%  & 0.9468 & 0.9796 & 0.9656 \\
    & 100 \% & 0.7298 & 0.6837 & 0.6866 \\
    & Overall     & 0.9471 & 0.9018 & 0.9215 \\
\midrule
\multirow{6}{*}{$\text{OOD}_m$}
    & 0 \%   & 0.9821 & 0.9965 & 0.9784 \\
    & 25 \%  & 0.8967 & 0.9402 & 0.8379 \\
    & 50 \%  & 0.9957 & 0.9971 & 0.9958 \\
    & 75 \%  & 0.8511 & 0.9002 & 0.8262 \\
    & 100 \% & 0.7933 & 0.6824 & 0.7212 \\
    & Overall     & 0.8768 & 0.8721 & 0.8638 \\
\midrule
\multirow{6}{*}{$\text{OOD}_t$}
    & 0 \%   & 1.0000 & 1.0000 & 1.0000 \\
    & 25 \%  & 0.9901 & 0.9909 & 0.9992 \\
    & 50 \%  & 0.9881 & 0.9594 & 0.9942 \\
    & 75 \%  & 0.8700 & 0.4276 & 0.7817 \\
    & 100 \% & 0.6050 & 0.2351 & 0.4262 \\
    & Overall     & 0.8712 & 0.7413 & 0.7142 \\
\bottomrule
\end{tabular}
}
\end{table}

%%%%%%%%%%%%%%%%%%%%%%%%%%%%%%%%%%%%%%%%%%%%%%%%%%%%%%%%%%%%%%%%

\begin{table}[pos=h!]
\centering
\caption{Ablation study of reconstruction components across targets. The Configuration tuple follows the order (Self, Shared, LOO) reconstruction, where \checkmark denotes inclusion and – denotes exclusion of the corresponding component}
\label{tab:ablation_ampere_combined}
\resizebox{\columnwidth}{!}{%
\begin{tabular}{l l c c c c c c}
\toprule
Setting & Configuration & 0\% & 25\% & 50\% & 75\% & 100\% & Overall \\
\midrule

\multirow{7}{*}{ID}
 & (\checkmark, \checkmark, \checkmark) & 1.0000 & 0.9991 & 0.9958 & 0.9468 & 0.7298 & 0.9471 \\
 & (–, \checkmark, \checkmark) & 1.0000 & 0.9909 & 0.9929 & 0.9459 & 0.7619 & 0.9412 \\
 & (\checkmark, –, \checkmark) & 0.9979 & 0.9893 & 0.9842 & 0.9447 & 0.7299 & 0.9216 \\
 & (\checkmark, \checkmark, –) & 0.5825 & 0.3206 & 0.5605 & 0.3939 & 0.4463 & 0.5018 \\
 & (\checkmark, –, –) & 0.1611 & 0.0554 & 0.0083 & 0.0270 & 0.1660 & 0.3413 \\
 & (–, \checkmark, –) & 0.4392 & 0.1179 & 0.2232 & 0.1081 & 0.2739 & 0.3846 \\
 & (–, –, \checkmark) & 0.9969 & 0.9823 & 0.9966 & 0.9456 & 0.7252 & 0.9094 \\

\midrule

\multirow{7}{*}{$\text{OOD}_m$}
 & (\checkmark, \checkmark, \checkmark) & 0.9821 & 0.8967 & 0.9957 & 0.8511 & 0.7933 & 0.8768 \\
 & (–, \checkmark, \checkmark) & 0.9739 & 0.8579 & 0.9956 & 0.8326 & 0.7995 & 0.8484 \\
 & (\checkmark, –, \checkmark) & 0.9427 & 0.8261 & 0.9939 & 0.7993 & 0.8036 & 0.8209 \\
 & (\checkmark, \checkmark, –) & 0.5669 & 0.2564 & 0.4834 & 0.3624 & 0.4329 & 0.4693 \\
 & (\checkmark, –, –) & 0.1342 & 0.0548 & 0.0070 & 0.0214 & 0.1707 & 0.3553 \\
 & (–, \checkmark, –) & 0.4352 & 0.1464 & 0.2090 & 0.0806 & 0.2921 & 0.3772 \\
 & (–, –, \checkmark) & 0.9123 & 0.8087 & 0.9880 & 0.7822 & 0.8036 & 0.8013 \\

\bottomrule
\end{tabular}
}
\end{table}

%%%%%%%%%%%%%%%%%%%%%%%%%%%%%%%%%%%%%%%%%%%%%%%%%%%%%%%%%%%%%%%%%%%%%%%%%%%%%%%%%%%%%%%%%%%%%%%%%%%%%%%%%%%%%%%%%%%%%%%%%%%%%%%%%%%%

\begin{table}[pos=h!]
\centering
\caption{Overall dynamic-threshold performance on the AMPERE dataset for different fault types and evaluation settings. *The configuration for the adaptive threshold has not been individually configured for these faults. }
\label{tab:ampere_overall_summary}
\resizebox{\columnwidth}{!}{%
\begin{tabular}{llcccc}
\toprule
Fault type & Setting & Accuracy & Precision & Recall & F1-score \\
\midrule
\multirow{3}{*}{One broken rotor bar}
& ID         & 0.9481 & 0.9652 & 0.9727 & 0.9689 \\  % UPDATED
& $\text{OOD}_m$ & 0.9335 & 0.9476 & 0.9516 & 0.9495 \\       % UPDATED
& $\text{OOD}_t$ & 0.9267 & 0.9584 & 0.9165 & 0.9370 \\       % UPDATED
\midrule
\multirow{3}{*}{Three broken rotor bars*}
& ID         & 0.9924 & 0.9909 & 1.0000 & 0.9954 \\
& $\text{OOD}_m$ & 0.9769 & 0.9661 & 1.0000 & 0.9828 \\
& $\text{OOD}_t$ & 0.9570 & 0.9327 & 1.0000 & 0.9652 \\
\midrule
\multirow{3}{*}{Four broken rotor bars*}
& ID         & 0.9924 & 0.9909 & 1.0000 & 0.9954 \\
& $\text{OOD}_m$ & 0.9769 & 0.9660 & 1.0000 & 0.9827 \\
& $\text{OOD}_t$ & 0.9569 & 0.9326 & 1.0000 & 0.9651 \\
\midrule
\multirow{3}{*}{Bearing degradation*}
& ID         & 0.9924 & 0.9910 & 1.0000 & 0.9955 \\
& $\text{OOD}_m$ & 0.9769 & 0.9662 & 1.0000 & 0.9828 \\
& $\text{OOD}_t$ & 0.9571 & 0.9329 & 1.0000 & 0.9653 \\
\bottomrule
\end{tabular}
}
\end{table}

%%%%%%%%%%%%%%%%%%%%%%%%%%%%%%%%%%%%%%%%%%%%%%%%%%%%%%%%%%%%%%%%
\paragraph{Evaluation Under Restructured Test-Stream Conditions}
%%%%%%%%%%%%%%%%%%%%%%%%%%%%%%%%%%%%%%%%%%%%%%%%%%%%%%%%%%%%%%%%
In previous experiments, the evaluation protocol assumed a sequential transition from exclusively healthy operation to faulty behavior. While convenient, this setup does not reflect realistic operating conditions, where faults may occur intermittently and under varying operational regimes.

To better approximate real-world scenarios, the test stream is restructured into a condition-aware sequence in which healthy and faulty samples are interleaved. In particular, for each operational condition (e.g., load levels such as 50\%, 75\%, and 100\%), the stream is initialized with healthy data before any faulty samples of the same condition are introduced. This ensures that the detection system observes a nominal baseline prior to fault exposure, while still encountering faults in a non-monotonic and temporally mixed manner.

\begin{figure}[pos = t]
    \centering
    \includegraphics[width=\linewidth]{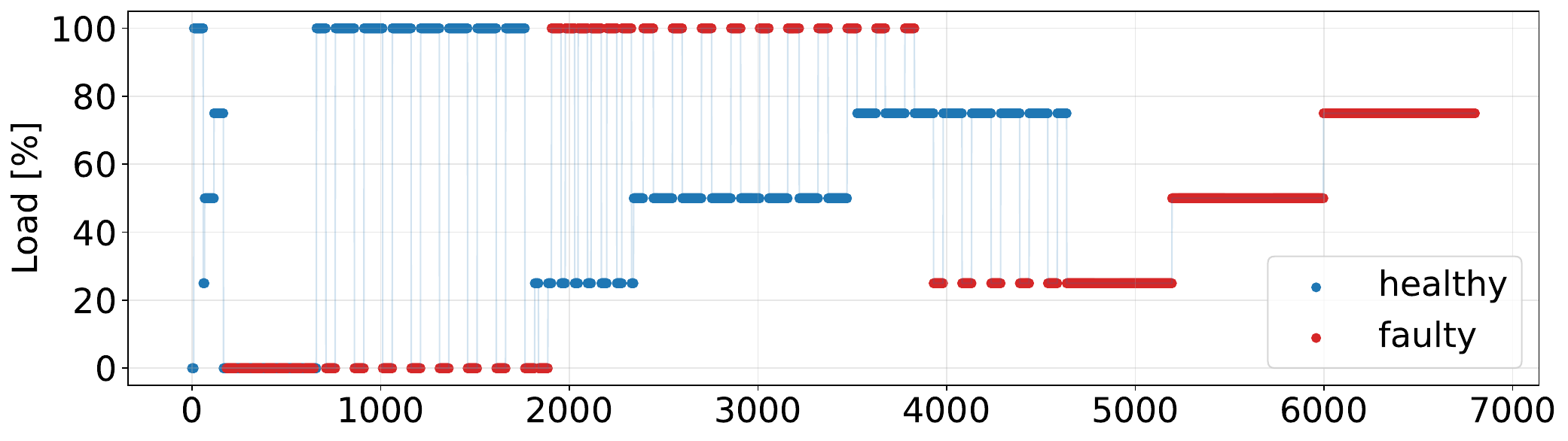}
    \caption{Illustration of the restructured test-stream used for evaluation. Healthy and faulty samples are interleaved across operating conditions, with each new condition initialized by healthy data before the introduction of faults. This operating condition emulates a more realistic, non-monotonic operating scenario compared to a purely sequential healthy-to-faulty transition. The x-axis is unit less, at the signal is artificial restructured.}    \label{fig:interleavedTestStream}
\end{figure}

The resulting evaluation stream, illustrated in Figure~\ref{fig:interleavedTestStream}, provides a more realistic and challenging setting for assessing detection performance, as it captures both distributional shifts across operating conditions and intermittent fault occurrences. We show the results, obtained on the Bearing Degradation problem, in Table~\ref{tab:interleavedTestStream}.

\begin{table}[pos = h]
\centering
\caption{Dynamic thresholding performance on the bearing degradation dataset under two evaluation protocols. The original test assumes a sequential transition from healthy to faulty operation, whereas the restructured test-stream interleaves healthy and faulty samples across operating conditions while ensuring condition-wise healthy initialization. Metrics are reported per condition and overall. The configuration for the adaptive threshold has not been individually configured for this test. }
\label{tab:interleavedTestStream}
\resizebox{\columnwidth}{!}{%
\begin{tabular}{l l cccc}
\toprule
Setting & Load  & Accuracy & Precision & Recall & F1-score \\
\midrule
\multirow{6}{*}{Restructured test-stream}
    & 0\%   & 0.9697 & 0.9653 & 1.0000 & 0.9823 \\
    & 25\%  & 0.9797 & 0.9759 & 1.0000 & 0.9878 \\
    & 50\%  & 0.9894 & 0.9793 & 1.0000 & 0.9895 \\
    & 75\%  & 0.9404 & 0.8932 & 1.0000 & 0.9436 \\
    & 100\% & 0.9750 & 0.9575 & 0.9952 & 0.9760 \\
    & Overall & 0.9702 & 0.9532 & 0.9990 & 0.9756 \\
\bottomrule
\end{tabular}
}
\end{table}

The results demonstrate that the proposed dynamic thresholding approach maintains robust detection performance under both evaluation protocols. In the original test setting, where the data follows a monotonic transition from healthy to faulty operation, the model achieves near-perfect recall across all conditions, with minor degradation in precision at higher load levels (75\% and 100\%).

Importantly, under the more realistic restructured test-stream-where healthy and faulty samples are interleaved across operating conditions-the performance remains stable and even improves slightly in terms of overall accuracy and precision. This indicates that the method generalizes well to non-stationary and temporally mixed scenarios, and is not reliant on an idealized ordering of the data.

Across all conditions, recall remains consistently at $0.995 - 1.000$, confirming that no faults are missed. The primary source of performance variation is precision, particularly at higher loads, suggesting that distributional shifts in operating conditions mainly affect false positive rates rather than detection sensitivity. Overall, these results validate the effectiveness of the proposed evaluation protocol and highlight the robustness of the model under realistic deployment conditions.

%%%%%%%%%%%%%%%%%%%%%%%%%%%%%%%%%%%%%%%%%%%%%%%%%%%%%%%%%%%%%%%%
\subsection{Results obtained on KAIST}\label{sc:AUCResults:KAIST}
%%%%%%%%%%%%%%%%%%%%%%%%%%%%%%%%%%%%%%%%%%%%%%%%%%%%%%%%%%%%%%%%
In Table~\ref{tab:KAISTAUC}, we report the AUC scores obtained with the proposed method, tested on the KAIST dataset. Table~\ref{tab:KAIST_overall_summary} summarizes the corresponding Accuracy, Precision, Recall, and F1-score using the adaptive thresholding strategy.

For the out-of-distribution settings, $\text{OOD}_m$ denotes that the $2\,\mathrm{Nm}$ load condition is excluded during training, while $\text{OOD}_t$ denotes that both $2\,\mathrm{Nm}$ and $4\,\mathrm{Nm}$ load conditions are excluded.

In addition, Table~\ref{tab:KAIST:ablation_components_combined} presents an ablation study evaluating the contribution of each component in the loss function. These results are obtained with less extensively tuned hyperparameters, as the Optuna-based optimization primarily targets the full model configuration.

\begin{table}[pos = h]
\centering
\caption{Performance comparison across settings.
AUC scores obtained with the proposed architecture on the \emph{KAIST} dataset, across operational parameters. Results obtained across all fault types. OOD meaning, 2\,Nm load is not present during training}
\label{tab:KAISTAUC}
\begin{tabular}{l l ccc}
\toprule
Setting & Load & Proposed method \\ %& Best Single Modality & Early stage sensor fusion\\
\midrule
\multirow{4}{*}{ID}
    & 0\,Nm   & 0.9555 \\%& 0.     & 0. \\
    & 2\,Nm   & 0.9993 \\%& 0.     & 0. \\
    & 4\,Nm   & 0.9993 \\%& 1.     & 0. \\
    & Overall & 0.9850 \\%& 0.     & 0. \\
\midrule
\multirow{4}{*}{$\text{OOD}_m$}
    & 0\,Nm   & 0.9531 \\%& 0.     & 0. \\
    & 2\,Nm   & 0.9876 \\%& 0.     & 0. \\
    & 4\,Nm   & 0.9983 \\%& 0.     & 0. \\
    & Overall & 0.9035 \\%& 0.     & 0. \\
\midrule
\multirow{4}{*}{$\text{OOD}_t$}
    & 0\,Nm   & 0.9739 \\%& 0.     & 0. \\
    & 2\,Nm   & 0.9735 \\%& 0.     & 0. \\
    & 4\,Nm   & 0.9983 \\%& 0.     & 0. \\
    & Overall & 0.8731 \\%& 0.     & 0. \\
    
\bottomrule
\end{tabular}%
\end{table}

%%%%%%%%%%%%%%%%%%%%%%%%%%%%%%%%%%%%%%%%%%%%%%%%%%%%%%%%%%%%%%%%%%%%%%%%%%%%%%%%%%%%%%%%%%%%%%%%%%%%%%%%%%%%%%%%%%%%%%%%%%%%%%%%%%%%

\begin{table}[pos=h]
\centering
\caption{Overall dynamic-threshold performance on the rotating machinery dataset across evaluation settings.}
\label{tab:KAIST_overall_summary}
\begin{tabular}{lcccc}
\toprule
Setting & Accuracy & Precision & Recall & F1-score \\
\midrule
ID                 & 0.9871 & 0.9880 & 0.9987 & 0.9933 \\
$\text{OOD}_m$            & 0.9759 & 0.9781 & 0.9965 & 0.9872 \\
$\text{OOD}_t$            & 0.9403 & 0.9405 & 0.9972 & 0.9680 \\
\bottomrule
\end{tabular}
\end{table}

%%%%%%%%%%%%%%%%%%%%%%%%%%%%%%%%%%%%%%%%%%%%%%%%%%%%%%%%%%%%%%%%%%%%%%%%%%%%%%%%%%%%%%%%%%%%%%%%%%%%%%%%%%%%%%%%%%%%%%%%%%%%%%%%%%%%

\begin{table}[pos=h!]
\centering
\caption{Ablation study of reconstruction components across settings. The Configuration tuple is ordered as (Self Reconstruction, Shared Reconstruction, Leave-One-Out Reconstruction), where \checkmark indicates the component is included and – indicates it is excluded.}
\label{tab:KAIST:ablation_components_combined}
\resizebox{\columnwidth}{!}{%

\begin{tabular}{l l c c c c}
\toprule
Setting & Configuration & 0\,Nm & 2\,Nm & 4\,Nm & Overall \\
\midrule

\multirow{7}{*}{ID}
 & (\checkmark, \checkmark, \checkmark) & 0.9503 & 0.9989 & 0.9990 & 0.9827 \\
 & (–, \checkmark, \checkmark) & 0.8910 & 0.9927 & 0.9984 & 0.9585 \\
 & (\checkmark, –, \checkmark) & 0.9312 & 0.9972 & 0.9992 & 0.9751 \\
 & (\checkmark, \checkmark, –) & 0.4515 & 0.9686 & 0.9994 & 0.7933 \\
 & (\checkmark, –, –) & 0.6066 & 0.8926 & 0.9993 & 0.8023 \\
 & (–, \checkmark, –) & 0.7716 & 0.8980 & 0.9991 & 0.8317 \\
 & (–, –, \checkmark) & 0.9467 & 0.9994 & 0.9997 & 0.9796 \\

\midrule

\multirow{7}{*}{$\text{OOD}_m$}
 & (\checkmark, \checkmark, \checkmark) & 0.9432 & 0.9568 & 0.9952 & 0.8930 \\
 & (–, \checkmark, \checkmark) & 0.9072 & 0.9264 & 0.9985 & 0.8645 \\
 & (\checkmark, –, \checkmark) & 0.9500 & 0.9405 & 0.9991 & 0.8894 \\
 & (\checkmark, \checkmark, –) & 0.6503 & 0.8924 & 0.9993 & 0.7612 \\
 & (\checkmark, –, –) & 0.7618 & 0.8866 & 0.9992 & 0.7714 \\
 & (–, \checkmark, –) & 0.8731 & 0.8727 & 0.9990 & 0.7844 \\
 & (–, –, \checkmark) & 0.9496 & 0.9384 & 0.9983 & 0.8793 \\

\bottomrule
\end{tabular}
}
\end{table}

%%%%%%%%%%%%%%%%%%%%%%%%%%%%%%%%%%%%%%%%%%%%%%%%%%%%%%%%%%%%%%%%
\subsection{Results obtained on IMAD-DS}\label{sc:AUCResults:IMAD}
%%%%%%%%%%%%%%%%%%%%%%%%%%%%%%%%%%%%%%%%%%%%%%%%%%%%%%%%%%%%%%%%
In Table~\ref{tab:IMADAUC}, we report the AUC scores obtained with the proposed method across the predefined source and target domains of the IMAD-DS dataset. In this case, the data are inherently partitioned based on operating speed, and we therefore adopt this split directly rather than using the ID/OOD terminology.

The target domain consists of speeds in the range 1000--1400 RPM, while the source domain comprises higher operating speeds. This setup reflects a distribution shift in operating conditions, where the target domain represents unseen regimes during training.

Table~\ref{tab:imad_overall} further reports Accuracy, Precision, Recall, and F1-score obtained using the proposed adaptive thresholding strategy.

\begin{table}[pos=h]
\centering
\caption{AUC scores obtained with the proposed architecture on the IMAD-DS dataset, across operational parameters. }
\label{tab:IMADAUC}
\begin{tabular}{l l cc}
\toprule
Domain & Speed (RPM) & AUC \\
\midrule
\multirow{5}{*}{Target}
    & 1000 & 0.8801 \\
    & 1100 & 0.9349 \\
    & 1200 & 0.9370 \\
    & 1300 & 0.9589 \\
    & 1400 & 0.9458 \\
\midrule
\multirow{9}{*}{Source}
    & 1500 & 0.9555 \\
    & 1600 & 0.9382 \\
    & 1700 & 0.9605 \\
    & 1800 & 0.9394 \\
    & 1900 & 0.8866 \\
    & 2000 & 0.8600 \\
    & 2400 & 0.9958 \\
    & 2800 & 0.9985 \\
    & 3000 & 1.0000 \\
\midrule
\multicolumn{2}{l}{Overall} & 0.6765 \\
\bottomrule
\end{tabular}
\end{table}

%%%%%%%%%%%%%%%%%%%%%%%%%%%%%%%%%%%%%%%%%%%%%%%%%%%%%%%%%%%%%%%%%%%%%%%%%%%%%%%%%%%%%%%%

\begin{table}[pos=h!]
\centering
\caption{Overall dynamic-threshold performance on IMAD-DS dataset.}
\label{tab:imad_overall}
\resizebox{\columnwidth}{!}{%
\begin{tabular}{lcccc}
\toprule
Setting & Accuracy & Precision & Recall & F1-score \\
\midrule
Overall                 & 0.896242 & 0.857548 & 0.950352 & 0.901568 \\
\bottomrule
\end{tabular}
}
\end{table}

%%%%%%%%%%%%%%%%%%%%%%%%%%%%%%%%%%%%%%%%%%%%%%%%%%%%%%%%%%%%%%%%
\section{Discussion}\label{sec:discussion} 
%%%%%%%%%%%%%%%%%%%%%%%%%%%%%%%%%%%%%%%%%%%%%%%%%%%%%%%%%%%%%%%%
This work introduced a multimodal reconstruction framework for anomaly detection and showed that cross-modal reconstruction improves robustness relative to unimodal and early-fusion baselines, particularly under severe out-of-distribution conditions. While performance differences are smaller in less challenging regimes, the proposed method degrades more gracefully as the mismatch between training and deployment conditions increases. The ablation study further indicates that this robustness is primarily driven by the leave-one-out reconstruction strategy, which encourages the model to learn shared latent structure rather than relying on modality-specific shortcuts.

Across all case studies, a degradation in AUC is observed under out-of-distribution (OOD) conditions, indicating that the learned representations are not fully domain invariant. Nevertheless, the proposed online, condition-aware threshold adaptation partially compensates for this degradation at the decision level and improves reliability under non-stationary operating conditions.
In the first case study, near-perfect performance is achieved under in-distribution conditions (F1 $\approx 0.99$ for $0$--$50\%$ load), and strong performance is retained under moderate OOD conditions ($75\%$ load). At extreme load ($100\%$), a noticeable degradation in both accuracy and F1-score is observed. Importantly, these results correspond to the most challenging fault scenario. For less severe fault cases, substantially higher robustness is observed, with F1-scores consistently exceeding $96\%$ and recall remaining at $1.00$, even under strong generalization. This indicates that the observed degradation is primarily attributable to intrinsic task difficulty under extreme operating conditions, rather than a fundamental limitation of the proposed method.
The degradation observed at the highest load level is consistent with prior classical broken-rotor-bar-diagnosis studies showing that fault indicators are strongly load dependent and may be confounded by load-related variations or oscillations, particularly for subtle rotor faults ~\cite{Ayhan2005,DIDIER20071127}. This suggests that, under extreme load, operating-condition effects can dominate the measured response and make reliable separation of normal and faulty behaviour more difficult.
In the second case study, the F1-score remains above $96\%$, while recall is consistently maintained at $98\%$, demonstrating that fault detection capability is largely preserved despite distributional shifts.
The third case study, achieved an overall F1 score of 0.90, lower than the other dataset - but also a different setup, where we include an acoustic measurement with artificial noise as part of the domain shift; not handled by the adaptive threshold. Even though lower, as evidenced by previously reported results, achieving an AUC of 0.5895 in the multimodal setting and 0.6930 using vibration signals only, the task remains highly challenging. 

\vspace{3mm}

Generally, the results suggest that robustness is influenced by the choice of sensing modalities, which differ across datasets. In particular, vibration-based signals introduce load-dependent variability that can obscure fault-specific patterns at high load levels, thereby increasing overlap between normal and faulty distributions, especially for subtle faults. This provides a plausible explanation for both the observed degradation under extreme conditions and discrepancies with prior work that excludes such modalities.

\vspace{3mm}

Overall, the results across all datasets demonstrate that dynamic, per-condition thresholding substantially improves robustness compared to static thresholding strategies, particularly in environments characterized by heterogeneous operating conditions and incomplete training coverage. By decoupling representation learning from decision calibration, combined with the multimodal cross-reconstruction architecture, the proposed approach enables reliable fault detection even when operational regimes encountered at inference time are not observed during training.

%%%%%%%%%%%%%%%%%%%%%%%%%%%%%%%%%%%%%%%%%%%%%%%%%%%%%%%%%%%%%%%%
\section{Conclusion}\label{sec:conclusion} 
%%%%%%%%%%%%%%%%%%%%%%%%%%%%%%%%%%%%%%%%%%%%%%%%%%%%%%%%%%%%%%%%
To conclude, we propose a multimodal reconstruction framework combined with condition-wise, gated exponentially weighted thresholding that adapts at test time. Across three datasets, the method achieves strong fault detection performance and, most importantly, demonstrates improved robustness under out-of-distribution operating conditions, particularly in the most challenging regimes where baseline methods degrade substantially.

Future work will focus on extending the approach to settings with continuously varying operational parameters, as the current formulation assumes discrete condition changes. In addition, the method relies on observable sources of domain shift; unmeasured factors, such as background noise or latent environmental variations, may degrade performance and require further investigation. It should further investigate optimal sensor selection, as this change the outcome of the methods capability. 

%%%%%%%%%%%%%%%%%%%%%%%%%%%%%%%%%%%%%%%%%%%%%%%%%%%%%%%%%
\section*{Acknowledgment of AI Assistance in Manuscript Preparation}
During the preparation of this work, the authors used Microsoft CoPilot to assist with refining and correcting the text. After using this tool, the authors carefully reviewed and edited the content as needed and take full responsibility for the content of this publication

\section*{Declaration of competing interest}
The authors declare that they have no known competing financial interests or personal relationships that could have appeared to influence the work reported in this paper

\section*{Acknowledgment of project funding}
Work by Magnus, Rafal and Dorte was supported by Innovation Fund Denmark under the BeWind project, grant 3148-00015A.

%%%%%%%%%%%%%%%%%%%%%%%%%%%%%%%%%%%%%%%%%%%%%%%%%%%%%%%%%%
%% Loading bibliography style file
%\bibliographystyle{model1-num-names}
\bibliographystyle{cas-model2-names}

% Loading bibliography database
\bibliography{cas-refs}

%%%%%%%%%%%%%%%%%%%%%%%%%%%%%%%%%%%%%%%%%%%%%%%%%%%%%%%%%%

%%%%%%%%%%%%%%%%%%%%%%%%%%%%%%%%%%%%%%%%%%%%%%%%%%%%%%%%%%%%%%%%
\appendix

%%%%%%%%%%%%%%%%%%%%%%%%%%%%%%%%%%%%%%%%%%%%%%%%%%%%%%%%%%%%%%%%%%%%%%%%%%
\FloatBarrier
\section{Detailed results for the AMPERE case study}\label{sc:app:detailedCaseStudy2}
%%%%%%%%%%%%%%%%%%%%%%%%%%%%%%%%%%%%%%%%%%%%%%%%%%%%%%%%%%%%%%%%%%%%%%%%%%

Tables \ref{tab:ampere_combined} to \ref{tab:ampere_bearing_strong_generalization} report detailed results for the AMPERE dataset using the adaptive threshold. For each experiment, we report Accuracy, Precision, Recall, and F1-score for all load conditions. The rows labeled In-distribution correspond to training and testing on the same fault case. The rows labeled $\text{OOD}_m$ correspond to the out-of-distribution setting where 25\% and 75\% is only at test time, and $\text{OOD}_t$ correspond to out-of-distribution settings where 50\%, 75\% and 100\% is only at test-time. Each table is organized by test fault type so that the evaluated condition is explicit.

\begin{table}[pos=h!]
\centering
\caption{AMPERE results with adaptive thresholding for the test fault case one broken rotor bar.}
\label{tab:ampere_combined}
\resizebox{\columnwidth}{!}{%
\begin{tabular}{l l cccc}
\toprule
Setting & Load & Accuracy & Precision & Recall & F1-score \\
\midrule

\multirow{6}{*}{In-distribution}
    & 0\%   & 0.9811 & 0.9781 & 1.0000 & 0.9889 \\
    & 25\%  & 0.9706 & 0.9664 & 1.0000 & 0.9829 \\
    & 50\%  & 0.9745 & 0.9701 & 1.0000 & 0.9848 \\
    & 75\%  & 0.9594 & 0.9527 & 1.0000 & 0.9758 \\
    & 100\% & 0.8546 & 0.9584 & 0.8628 & 0.9081 \\
    & Overall & 0.9481 & 0.9652 & 0.9727 & 0.9689 \\
\midrule

\multirow{6}{*}{$\text{OOD}_m$}
    & 0\%   & 0.9780 & 0.9745 & 1.0000 & 0.9871 \\
    & 25\%  & 0.9472 & 0.9110 & 0.9913 & 0.9494 \\
    & 50\%  & 0.9776 & 0.9736 & 1.0000 & 0.9866 \\
    & 75\%  & 0.9535 & 0.9148 & 1.0000 & 0.9555 \\
    & 100\% & 0.7875 & 0.9746 & 0.7656 & 0.8575 \\
    & Overall & 0.9335 & 0.9476 & 0.9516 & 0.9495 \\
\midrule

\multirow{6}{*}{$\text{OOD}_t$}
    & 0\%   & 0.9937 & 0.9926 & 1.0000 & 0.9963 \\
    & 25\%  & 0.9795 & 0.9758 & 1.0000 & 0.9877 \\
    & 50\%  & 0.9827 & 0.9666 & 1.0000 & 0.9830 \\
    & 75\%  & 0.9492 & 0.9076 & 1.0000 & 0.9515 \\
    & 100\% & 0.7756 & 0.9510 & 0.5810 & 0.7214 \\
    & Overall & 0.9267 & 0.9584 & 0.9165 & 0.9370 \\

\bottomrule
\end{tabular}
}
\end{table}

%%%%%%%%%%%%%%%%%%%%%%%%%%%%%%%%%%%%%%
\begin{table}[pos=h!]
\centering
\caption{AMPERE results with adaptive thresholding for the test fault case three broken rotor bars. The adaptive-threshold configuration was not tuned separately for this fault type.}
\label{tab:ampere_strong_generalization}
\resizebox{\columnwidth}{!}{%
\begin{tabular}{l l cccc}
\toprule
Setting & Load & Accuracy & Precision & Recall & F1-score \\
\midrule
\multirow{6}{*}{In-distribution}
    & 0\%  & 0.9958 & 0.9951 & 1.0000 & 0.9975 \\
    & 25\% & 0.9906 & 0.9890 & 1.0000 & 0.9945 \\
    & 50\% & 0.9928 & 0.9914 & 1.0000 & 0.9957 \\
    & 75\% & 0.9909 & 0.9890 & 1.0000 & 0.9945 \\
    & 100\%& 0.9918 & 0.9902 & 1.0000 & 0.9951 \\
    & Overall & 0.9924 & 0.9909 & 1.0000 & 0.9954 \\
\midrule

\multirow{6}{*}{$\text{OOD}_m$}
    & 0\%  & 0.9948 & 0.9938 & 1.0000 & 0.9969 \\
    & 25\% & 0.9351 & 0.8855 & 1.0000 & 0.9393 \\
    & 50\% & 0.9949 & 0.9938 & 1.0000 & 0.9969 \\
    & 75\% & 0.9846 & 0.9701 & 1.0000 & 0.9848 \\
    & 100\%& 0.9979 & 0.9975 & 1.0000 & 0.9988 \\
    & Overall & 0.9769 & 0.9661 & 1.0000 & 0.9828 \\
\midrule

\multirow{6}{*}{$\text{OOD}_t$}
    & 0\%  & 1.0000 & 1.0000 & 1.0000 & 1.0000 \\
    & 25\% & 0.9939 & 0.9927 & 1.0000 & 0.9963 \\
    & 50\% & 0.9926 & 0.9853 & 1.0000 & 0.9926 \\
    & 75\% & 0.9259 & 0.8712 & 1.0000 & 0.9312 \\
    & 100\%& 0.9046 & 0.8406 & 1.0000 & 0.9134 \\
    & Overall & 0.9570 & 0.9327 & 1.0000 & 0.9652 \\

\bottomrule
\end{tabular}
}
\end{table}

%%%%%%%%%%%%%%%%%%%%%%%%%%%%%%%%%%%%%%%%%%%%%%%%%%%%%%%%%%%%%%%%%%%%
\begin{table}[pos=h!]
\centering
\caption{AMPERE results with adaptive thresholding for the test fault case four broken rotor bars. The adaptive-threshold configuration was not tuned separately for this fault type.}
\label{tab:ampere_four_strong_generalization}
\resizebox{\columnwidth}{!}{%
\begin{tabular}{l l cccc}
\toprule
Setting & Load & Accuracy & Precision & Recall & F1-score \\
\midrule
\multirow{6}{*}{In-distribution}
    & 0\%  & 0.9958 & 0.9951 & 1.0000 & 0.9975 \\
    & 25\% & 0.9906 & 0.9890 & 1.0000 & 0.9945 \\
    & 50\% & 0.9929 & 0.9915 & 1.0000 & 0.9957 \\
    & 75\% & 0.9909 & 0.9890 & 1.0000 & 0.9944 \\
    & 100\%& 0.9917 & 0.9902 & 1.0000 & 0.9951 \\
    & Overall & 0.9924 & 0.9909 & 1.0000 & 0.9954 \\
\midrule

\multirow{6}{*}{$\text{OOD}_m$}
    & 0\%  & 0.9948 & 0.9939 & 1.0000 & 0.9969 \\
    & 25\% & 0.9349 & 0.8849 & 1.0000 & 0.9389 \\
    & 50\% & 0.9949 & 0.9939 & 1.0000 & 0.9969 \\
    & 75\% & 0.9845 & 0.9699 & 1.0000 & 0.9847 \\
    & 100\%& 0.9979 & 0.9975 & 1.0000 & 0.9988 \\
    & Overall & 0.9769 & 0.9660 & 1.0000 & 0.9827 \\
\midrule

\multirow{6}{*}{$\text{OOD}_t$}
    & 0\%  & 1.0000 & 1.0000 & 1.0000 & 1.0000 \\
    & 25\% & 0.9939 & 0.9926 & 1.0000 & 0.9963 \\
    & 50\% & 0.9926 & 0.9854 & 1.0000 & 0.9927 \\
    & 75\% & 0.9257 & 0.8704 & 1.0000 & 0.9307 \\
    & 100\%& 0.9042 & 0.8396 & 1.0000 & 0.9128 \\
    & Overall & 0.9569 & 0.9326 & 1.0000 & 0.9651 \\
\bottomrule
\end{tabular}
}
\end{table}

%%%%%%%%%%%%%%%%%%%%%%%%%%%%%%%%%%%%%%%%%%%%%%%%%%%%%%%%%%%%%%%%%%%%
\begin{table}[pos=h!]
\centering
\caption{AMPERE results with adaptive thresholding for the test fault case bearing degradation. The adaptive-threshold configuration was not tuned separately for this fault type.}
\label{tab:ampere_bearing_strong_generalization}
\resizebox{\columnwidth}{!}{%
\begin{tabular}{l l cccc}
\toprule
Setting & Load & Accuracy & Precision & Recall & F1-score \\
\midrule
\multirow{6}{*}{In-distribution}
    & 0\%  & 0.9958 & 0.9951 & 1.0000 & 0.9975 \\
    & 25\% & 0.9906 & 0.9890 & 1.0000 & 0.9945 \\
    & 50\% & 0.9928 & 0.9914 & 1.0000 & 0.9957 \\
    & 75\% & 0.9908 & 0.9889 & 1.0000 & 0.9944 \\
    & 100\%& 0.9920 & 0.9905 & 1.0000 & 0.9952 \\
    & Overall & 0.9924 & 0.9910 & 1.0000 & 0.9955 \\
\midrule

\multirow{6}{*}{$\text{OOD}_m$}
    & 0\%  & 0.9948 & 0.9938 & 1.0000 & 0.9969 \\
    & 25\% & 0.9351 & 0.8854 & 1.0000 & 0.9392 \\
    & 50\% & 0.9949 & 0.9938 & 1.0000 & 0.9969 \\
    & 75\% & 0.9845 & 0.9698 & 1.0000 & 0.9847 \\
    & 100\%& 0.9980 & 0.9976 & 1.0000 & 0.9988 \\
    & Overall & 0.9769 & 0.9662 & 1.0000 & 0.9828 \\
\midrule

\multirow{6}{*}{$\text{OOD}_t$}
    & 0\%  & 1.0000 & 1.0000 & 1.0000 & 1.0000 \\
    & 25\% & 0.9939 & 0.9927 & 1.0000 & 0.9963 \\
    & 50\% & 0.9925 & 0.9853 & 1.0000 & 0.9926 \\
    & 75\% & 0.9255 & 0.8700 & 1.0000 & 0.9305 \\
    & 100\%& 0.9060 & 0.8446 & 1.0000 & 0.9158 \\
    & Overall & 0.9571 & 0.9329 & 1.0000 & 0.9653 \\
\bottomrule
\end{tabular}
}
\end{table}

%%%%%%%%%%%%%%%%%%%%%%%%%%%%%%%%%%%%%%%%%%%%%%%%%%%%%%%%%%%%%%%%
\FloatBarrier
\section{Detailed results for the KAIST case study}\label{sc:app:detailedCaseStudy1}
%%%%%%%%%%%%%%%%%%%%%%%%%%%%%%%%%%%%%%%%%%%%%%%%%%%%%%%%%%%%%%%%%%%%%%%%%%

Table \ref{tab:ood_dynamic_threshold_combined} reports detailed results for the KAIST dataset using the adaptive threshold and we report Accuracy, Precision, Recall, and F1-score for each load condition. The rows labeled In-distribution correspond to training and testing on the same opertaional conditions. The rows labeled $\text{OOD}_m$ correspond to the out-of-distribution setting where 2\,Nm is only at test time, and $\text{OOD}_t$ correspond to out-of-distribution settings where 2\,Nm and 4\,Nm is only at test-time.

\begin{table}[pos=h!]
\centering
\caption{KAIST results with adaptive thresholding across in-distribution, simple-generalization, and strong-generalization settings.}
\label{tab:ood_dynamic_threshold_combined}
\resizebox{\columnwidth}{!}{%
\begin{tabular}{l l cccc}
\toprule
Setting & Load & Accuracy & Precision & Recall & F1-score \\
\midrule

\multirow{4}{*}{In-distribution}
    & 0\,Nm            & 0.9785 & 0.9807 & 0.9970 & 0.9888 \\
    & 2\,Nm            & 0.9887 & 0.9883 & 1.0000 & 0.9941 \\
    & 4\,Nm            & 0.9951 & 0.9952 & 0.9998 & 0.9975 \\
    & Overall & 0.9871 & 0.9880 & 0.9987 & 0.9933 \\
\midrule

\multirow{4}{*}{$\text{OOD}_m$}
    & 0\,Nm            & 0.9777 & 0.9774 & 0.9996 & 0.9884 \\
    & 2\,Nm            & 0.9406 & 0.9465 & 0.9830 & 0.9644 \\
    & 4\,Nm            & 0.9940 & 0.9942 & 0.9997 & 0.9969 \\
    & Overall & 0.9759 & 0.9781 & 0.9965 & 0.9872 \\
\midrule

\multirow{4}{*}{$\text{OOD}_t$}
    & 0\,Nm            & 0.9798 & 0.9792 & 1.0000 & 0.9895 \\
    & 2\,Nm            & 0.8333 & 0.8337 & 0.9946 & 0.9071 \\
    & 4\,Nm            & 0.9588 & 0.9605 & 0.9954 & 0.9776 \\
    & Overall & 0.9404 & 0.9405 & 0.9971 & 0.9680 \\
\bottomrule
\end{tabular}
}
\end{table}

%%%%%%%%%%%%%%%%%%%%%%%%%%%%%%%%%%%%%%%%%%%%%%%%%%%%%%%%%%
\FloatBarrier
\section{Detailed results for the IMAD-DS case study}\label{sc:app:castStudy3}
%%%%%%%%%%%%%%%%%%%%%%%%%%%%%%%%%%%%%%%%%%%%%%%%%%%%%%%%%%

Table \ref{tab:IMAD_DETAILED} reports detailed results for the IMAD-DS dataset using the adaptive threshold. The table is grouped by domain so that the distinction between target-domain and source-domain operating conditions is explicit. Accuracy, Precision, Recall, and F1-score are reported for each rotational speed.

\begin{table}[pos=h!]
\centering
\caption{IMAD-DS results with adaptive thresholding across target-domain and source-domain operating conditions.}
\label{tab:IMAD_DETAILED}
\resizebox{\columnwidth}{!}{%
\begin{tabular}{l l cccc}
\toprule
Domain & Speed (RPM) & Accuracy & Precision & Recall & F1-score \\
\midrule
\multirow{5}{*}{Target domain}
 & 1000 & 0.8756 & 0.9028 & 0.8419 & 0.8713 \\
 & 1100 & 0.9247 & 0.9224 & 0.9326 & 0.9275 \\
 & 1200 & 0.9163 & 0.9057 & 0.9294 & 0.9174 \\
 & 1300 & 0.9283 & 0.9216 & 0.9363 & 0.9289 \\
 & 1400 & 0.9197 & 0.8753 & 0.9725 & 0.9214 \\
\midrule
\multirow{9}{*}{Source domain}
 & 1500 & 0.9008 & 0.8447 & 0.9956 & 0.9140 \\
 & 1600 & 0.8787 & 0.8121 & 0.9853 & 0.8904 \\
 & 1700 & 0.8617 & 0.7852 & 0.9956 & 0.8780 \\
 & 1800 & 0.8812 & 0.8020 & 0.9926 & 0.8872 \\
 & 1900 & 0.8704 & 0.8448 & 0.9392 & 0.8895 \\
 & 2000 & 0.8189 & 0.7588 & 0.9020 & 0.8242 \\
 & 2400 & 0.9020 & 0.8361 & 1.0000 & 0.9107 \\
 & 2800 & 0.9052 & 0.8407 & 1.0000 & 0.9134 \\
 & 3000 & 0.8973 & 0.8209 & 1.0000 & 0.9017 \\
\midrule
\multicolumn{2}{l}{Overall} & 0.8912 & 0.8519 & 0.9583 & 0.9013 \\
\bottomrule
\end{tabular}
}
\end{table}

%%%%%%%%%%%%%%%%%%%%%%%%%%%%%%%%%%%%%%%%%%%%%%%%%%%%%%%%%%
\FloatBarrier
\section{Hyperparameters}\label{sc:app:Hyperparameters}
%%%%%%%%%%%%%%%%%%%%%%%%%%%%%%%%%%%%%%%%%%%%%%%%%%%%%%%%%%

The model configurations used in the experiments are summarized in Tables \ref{tab:ampere_id_ood} to \ref{tab:hyperparameters_imadds}. For consistency with the main text, ID denotes the in-distribution setting, $\text{OOD}_m$ denotes the simple-generalization setting, and $\text{OOD}_t$ denotes the strong-generalization setting. Table \ref{tab:ampere_id_ood} reports the proposed model for AMPERE with the one-broken-rotor-bar setup, Table \ref{tab:baselineHyperparameters} reports the AMPERE baselines, Table \ref{tab:kaist_id_ood_simple} reports the KAIST configuration, and Table \ref{tab:hyperparameters_imadds} reports the IMAD-DS configuration.

%%%%%%%%%%%%%%%%%%%%%%%%%%%%%%%%%%%%%%%%%%%%%%%%%%%%%%%%%%%%%%%%%%%%%%
\begin{table}[pos = h]
\centering
\caption{Hyperparameter configuration for the proposed model on the AMPERE dataset under ID, $\text{OOD}_m$, and $\text{OOD}_t$ settings, using the one-broken-rotor-bar setup.}
\label{tab:ampere_id_ood}
\resizebox{\columnwidth}{!}{%
\begin{tabular}{lccc}
\toprule
Category & ID & $\text{OOD}_m$ & $\text{OOD}_t$ \\
\midrule

\multicolumn{4}{l}{Latent space and attention} \\
Latent dimension (per modality) & $64$ & $64$ & $64$ \\
Attention heads & $1$ & $1$ & $2$ \\

\midrule
\multicolumn{4}{l}{Encoder architecture} \\
Voltage encoder & $512 \rightarrow 512$ & $512 \rightarrow 512$ & $2560 \rightarrow 1024$ \\
Current encoder & $2048 \rightarrow 512$ & $2048 \rightarrow 512$ & $512 \rightarrow 128$ \\
Acceleration encoder & $1024 \rightarrow 512$ & $1024 \rightarrow 512$ & $2560 \rightarrow 128$ \\

\midrule
\multicolumn{4}{l}{Loss weighting coefficients} \\
Self-reconstruction & $0.226$ & $0.226$ & $0.699$ \\
Shared-reconstruction & $0.978$ & $0.978$ & $0.444$ \\
Leave-one-out reconstruction & $1.382$ & $1.382$ & $2.140$ \\

\midrule
\multicolumn{4}{l}{Optimization} \\
Learning rate & $8.45 \times 10^{-4}$ & $8.45 \times 10^{-4}$ & $2.82 \times 10^{-4}$ \\
Weight decay & $6.09 \times 10^{-7}$ & $6.09 \times 10^{-7}$ & $7.57 \times 10^{-6}$ \\
Batch size & $64$ & $64$ & $64$ \\

\midrule
\multicolumn{4}{l}{Training schedule} \\
Epochs & $50$ & $50$ & $50$ \\
Early stopping patience & $5$ & $5$ & $5$ \\
Warm-up epochs & $0$ & $0$ & $0$ \\

\midrule
\multicolumn{4}{l}{Adaptive thresholding and stability} \\
Alarm threshold             & $3.5\sigma$ & $3\sigma$ & $3.5\sigma$ \\
OOD threshold               & $4.0\sigma$ & $7\sigma$ & $7\sigma$ \\
Update threshold            & $0.1\sigma$ & $0.1\sigma$ & $1.5\sigma$ \\
Adaptation rate ($\alpha$)  & $0.1$ & $0.05$ & $0.1$ \\
Min. std. deviation         & $10^{-6}$ & $10^{-6}$ & $10^{-6}$ \\

\bottomrule
\end{tabular}
}
\end{table}

%%%%%%%%%%%%%%%%%%%%%%%%%%%%%%%%%%%%%%%%%%%%%%%%%%%%%
\begin{table}[pos = h]
\centering
\caption{Hyperparameter configurations for the AMPERE baselines under ID, $\text{OOD}_m$, and $\text{OOD}_t$ settings. The two baseline models are the unimodal acceleration autoencoder and the early-fusion multimodal model.}
\label{tab:baselineHyperparameters}
\resizebox{\columnwidth}{!}{%
\begin{tabular}{lccc}
\toprule
Category & ID & $\text{OOD}_m$ & $\text{OOD}_t$ \\
\midrule

\multicolumn{4}{l}{Unimodal acceleration autoencoder} \\
Latent dimension & $32$ & $128$ & $16$ \\
Encoder architecture & $1024 \rightarrow 512$ & $256 \rightarrow 256$ & $1024 \rightarrow 512$ \\
Learning rate & $1.88 \times 10^{-4}$ & $1.02 \times 10^{-3}$ & $1.45 \times 10^{-3}$ \\
Weight decay & $1.37 \times 10^{-5}$ & $2.91 \times 10^{-8}$ & $1.91 \times 10^{-4}$ \\
Batch size & $64$ & $32$ & $128$ \\
Epochs & $50$ & $50$ & $50$ \\
Early stopping patience & $5$ & $5$ & $5$ \\
Warm-up epochs & $0$ & $0$ & $0$ \\

\midrule
\multicolumn{4}{l}{Early-fusion multimodal model} \\
Latent dimension & $256$ & $64$ & $256$ \\
Encoder architecture & $512 \rightarrow 512$ & $4096 \rightarrow 4096$ & $2048 \rightarrow 1024$ \\
Learning rate & $5.73 \times 10^{-4}$ & $1.04 \times 10^{-4}$ & $1.11 \times 10^{-4}$ \\
Weight decay & $1.07 \times 10^{-7}$ & $1.16 \times 10^{-7}$ & $2.04 \times 10^{-8}$ \\
Batch size & $32$ & $64$ & $64$ \\
Epochs & $50$ & $50$ & $50$ \\
Early stopping patience & $5$ & $5$ & $5$ \\
Warm-up epochs & $0$ & $0$ & $0$ \\

\bottomrule
\end{tabular}
}
\end{table}

%%%%%%%%%%%%%%%%%%%%%%%%%%%%%%%%%%%%%%%%%%%%%%%%%%%%%%

\begin{table}[pos = h]
\centering
\caption{Hyperparameter configuration for the proposed model on the KAIST dataset under ID, $\text{OOD}_m$, and $\text{OOD}_t$ settings.}
\label{tab:kaist_id_ood_simple}
\resizebox{\columnwidth}{!}{%
\begin{tabular}{lccc}
\toprule
Category & ID & $\text{OOD}_m$ & $\text{OOD}_t$ \\
\midrule

\multicolumn{4}{l}{Latent space and attention} \\
Latent dimension (per modality) & $16$ & $16$ & $32$ \\
Attention heads & $2$ & $2$ & $2$ \\

\midrule
\multicolumn{4}{l}{Encoder architecture} \\
Vibration encoder & $2560 \rightarrow 1902$ & $2560 \rightarrow 1902$ & $2048 \rightarrow 1024$ \\
Temperature encoder & $1024 \rightarrow 1024$ & $1024 \rightarrow 1024$ & $256 \rightarrow 128$ \\
Current encoder & $1024 \rightarrow 512$ & $1024 \rightarrow 512$ & $256 \rightarrow 256$ \\

\midrule
\multicolumn{4}{l}{Loss weighting coefficients} \\
Self-reconstruction & $0.186$ & $0.186$ & $0.825$ \\
Shared-reconstruction & $0.703$ & $0.703$ & $0.270$ \\
Leave-one-out reconstruction & $2.213$ & $2.213$ & $2.685$ \\

\midrule
\multicolumn{4}{l}{Optimization} \\
Learning rate & $1.73 \times 10^{-4}$ & $1.73 \times 10^{-4}$ & $5.92 \times 10^{-4}$ \\
Weight decay & $1.46 \times 10^{-7}$ & $1.46 \times 10^{-7}$ & $8.43 \times 10^{-6}$ \\
Batch size & $128$ & $128$ & $64$ \\

\midrule
\multicolumn{4}{l}{Training schedule} \\
Epochs & $50$ & $50$ & $50$ \\
Early stopping patience & $5$ & $5$ & $5$ \\
Warm-up epochs & $0$ & $0$ & $0$ \\

\midrule
\multicolumn{4}{l}{Adaptive thresholding and stability} \\
Alarm threshold & $3.5\sigma$ & $3.5\sigma$ & $3.5\sigma$ \\
OOD threshold & $4.0\sigma$ & $7.0\sigma$ & $7.0\sigma$ \\
Update threshold & $1.5\sigma$ & $0.1\sigma$ & $2.0\sigma$ \\
Adaptation rate ($\alpha$) & $0.05$ & $0.05$ & $0.1$ \\
Min. std. deviation & $10^{-6}$ & $10^{-6}$ & $10^{-6}$ \\

\bottomrule
\end{tabular}
}
\end{table}

%%%%%%%%%%%%%%%%%%%%%%%%%%%%%%%%%%%%%%%%%%%%%%%%%%%%%%%%%%%%%%%%%%%%%%
\begin{table}[pos = h]
\centering
\caption{Hyperparameter configuration for the proposed model on the IMAD-DS dataset.}
\label{tab:hyperparameters_imadds}
\begin{tabular}{lc}
\toprule
Category & IMAD-DS \\
\midrule

\multicolumn{2}{l}{Latent space and attention} \\
Latent dimension & $64$ \\
Attention heads & $2$ \\

\midrule
\multicolumn{2}{l}{Encoder architecture} \\
Acceleration encoder & $2048 \rightarrow 256$ \\
Microphone encoder & $2048 \rightarrow 64$ \\
Gyroscope encoder & $2048 \rightarrow 128$ \\

\midrule
\multicolumn{2}{l}{Loss weighting coefficients} \\
Self-reconstruction & $0.980$ \\
Shared-reconstruction & $0.259$ \\
Leave-one-out reconstruction & $3.753$ \\

\midrule
\multicolumn{2}{l}{Optimization} \\
Learning rate & $1.02 \times 10^{-4}$ \\
Weight decay & $1.31 \times 10^{-6}$ \\
Batch size & $128$ \\

\midrule
\multicolumn{2}{l}{Training schedule} \\
Epochs & $500$ \\
Early stopping patience & $5$ \\
Warm-up epochs & $0$ \\

\midrule
\multicolumn{2}{l}{Adaptive thresholding and stability} \\
Alarm threshold & $3.5\sigma$ \\
OOD threshold & $4.5\sigma$ \\
Update threshold & $0.1\sigma$ \\
Adaptation rate ($\alpha$) & $0.05$ \\
Min. std. deviation & $10^{-6}$ \\

\bottomrule
\end{tabular}
\end{table}

% To print the credit authorship contribution details
\printcredits

% Biography
%\bio{}
% Here goes the biography details.
%\endbio

%\bio{pic1}
% Here goes the biography details.
%\endbio

\end{document}